\pdfoutput=1 

\documentclass[sigconf,nonacm]{aamas}   

\usepackage{balance} 

\usepackage{booktabs}
\usepackage{tabularx}
\usepackage{array}
\usepackage{graphicx}
\usepackage{pifont}
\usepackage{threeparttable}
\usepackage{multirow}
\usepackage{makecell}
\usepackage{wrapfig}

\usepackage{algorithm}

\usepackage[most]{tcolorbox}
\usepackage{listings}
\usepackage{algorithm}
\usepackage{xcolor}

\usepackage{soul}
\usepackage{xcolor}

\definecolor{pbBlue}{HTML}{1F4E79}
\definecolor{pbTeal}{HTML}{00695C}
\definecolor{pbOrange}{HTML}{D35400}
\definecolor{pbGreen}{HTML}{2E7D32}
\definecolor{pbRed}{HTML}{B71C1C}
\definecolor{pbGray}{HTML}{546E7A}

\usepackage{etoolbox}

\makeatletter
\patchcmd{\@mkauthors@iii}
  {\par\bigskip}
  {\par\bigskip\vspace{6pt}}
  {}
  {\PackageWarning{main_supp}{Author spacing patch failed}}
\makeatother

\lstdefinestyle{promptstyle}{
  basicstyle=\ttfamily\scriptsize, columns=fullflexible, keepspaces=true,
  breaklines=true, breakatwhitespace=true, showstringspaces=false, upquote=true,
  breakindent=1.2em, postbreak=\mbox{\textcolor{pbGray}{$\hookrightarrow$}\space},
  moredelim=[l][\color{pbBlue}\bfseries]{\#\#},
  emph={FINAL,ACTIONS,BLOCKS,BLOCK,END,NO_ACTIONS}, emphstyle=\color{pbOrange}\bfseries,
  emph={[2]Preconditions,Effects,urgency,insert,objects,reason}, emphstyle={[2]\color{pbTeal}\bfseries},
}

\newtcblisting{promptbox}[3][]{
  enhanced, listing only, listing options={style=promptstyle},
  colback=#2!3!white, colframe=#2!85!black, boxrule=0.5pt, arc=1.5mm,
  left=1.5mm, right=1.5mm, top=2.5mm, bottom=0.5mm, before skip=2.5mm, after skip=1mm,
  fonttitle=\sffamily\bfseries\footnotesize, coltitle=white, title={#3},
  attach boxed title to top left={yshift=-2.2mm, xshift=3mm},
  boxed title style={colback=#2!85!black, arc=1mm, boxrule=0pt, left=1.5mm, right=1.5mm},
  #1}

\usepackage{graphicx}
\usepackage{booktabs}
\usepackage{algorithm}
\usepackage{multirow}
\usepackage{array}
\usepackage{tabularx}

\usepackage{multirow}

\usepackage{color,xcolor,ucs}

\usepackage{algorithm}
\usepackage[noend]{algpseudocode}

\usepackage{caption}

\newif\ifsuppcombined
\suppcombinedtrue

\newcommand{\institutionsfootnote}{%
\begingroup
\renewcommand{\thefootnotecopyrightpermission}{}%
\footnotetextcopyrightpermission{%
\footnotesize\noindent
  \textsuperscript{1}Robotics Research Center, IIIT Hyderabad, India.\\
  \textsuperscript{2}Center for Security, Theory and Algorithmic Research, IIIT Hyderabad, India.\\
  \textsuperscript{$\dagger$}All authors contributed equally to this work.\\
  \textsuperscript{$\ast$}Corresponding author: \texttt{narendhiran.v@research.iiit.ac.in}
}%
\endgroup
}

\copyrightyear{2026}
\acmYear{2026}
\acmDOI{}
\acmISBN{}

\acmSubmissionID{2220}

\submissionType{}

\title{Deciding If, When, and Where to Replan for Robust Manipulation}

\author{
Narendhiran Vijayakumar\textsuperscript{1\char44\ensuremath{\dagger}\char44\ensuremath{\ast}},
Puru Ojha\textsuperscript{1\char44\ensuremath{\dagger}},
Nav Singhal\textsuperscript{1\char44\ensuremath{\dagger}},
Girish Varma\textsuperscript{2},
Antony Thomas\textsuperscript{1}
}

\renewcommand{\shortauthors}{Vijayakumar et al.}

\begin{abstract}

Foundation-model (FM) based task and motion planning (TAMP) methods typically assume that the full scene is known before planning. In long-horizon manipulation, however, relevant objects may initially be hidden and only become observable as the robot interacts with the environment. We introduce \textsc{Robust-TAMP}, a task and motion planning framework that first decides whether a discovery needs a replan. If replanning is needed, the robot executes unaffected actions while the FM synthesizes corrective actions. These actions are inserted into the existing plan based on urgency, either executed immediately or deferred while preserving the remaining plan. We evaluate \textsc{Robust-TAMP} on 14 MuJoCo environments with hidden objects, blocked placements, articulated containers, and multi-step tasks, and on five real-world settings. Our proposed method completes 92.1\% of simulated tasks, outperforming four baseline comparison architectures by 47.1 percentage points under the same settings. Ablations show that each component either raises success or lowers planner calls, planning time, and robot idle time. An extensive study across model sizes, text and vision inputs, reasoning modes, and prompting further shows that in-context examples help the planner follow multi-step procedures that zero-shot prompting struggles with.  Our implementation code and project details are available at \url{https://narendhiranv04.github.io/ROBUST-TAMP/}.
\end{abstract}

\newcolumntype{Y}{>{\raggedright\arraybackslash}X}

\newcolumntype{L}[1]{>{\raggedright\arraybackslash}p{#1}}

\definecolor{mygray}{gray}{0.6}

\usepackage{xcolor}

\usepackage[normalem]{ulem}

\begin{document}



\maketitle

\institutionsfootnote
\fancypagestyle{standardpagestyle}{\fancyhf{}\renewcommand{\headrulewidth}{0pt}\fancyfoot[C]{\footnotesize\thepage}}
\fancypagestyle{firstpagestyle}{\fancyhf{}\renewcommand{\headrulewidth}{0pt}\fancyfoot[C]{\footnotesize\thepage}}
\thispagestyle{firstpagestyle}\pagestyle{standardpagestyle}


\section{Introduction}
\label{sec:intro}


\begin{figure}[t]
    \centering
    \includegraphics[
        width=0.98\columnwidth,
        trim={0.25cm 2.05cm 2cm 0cm},
        clip
    ]{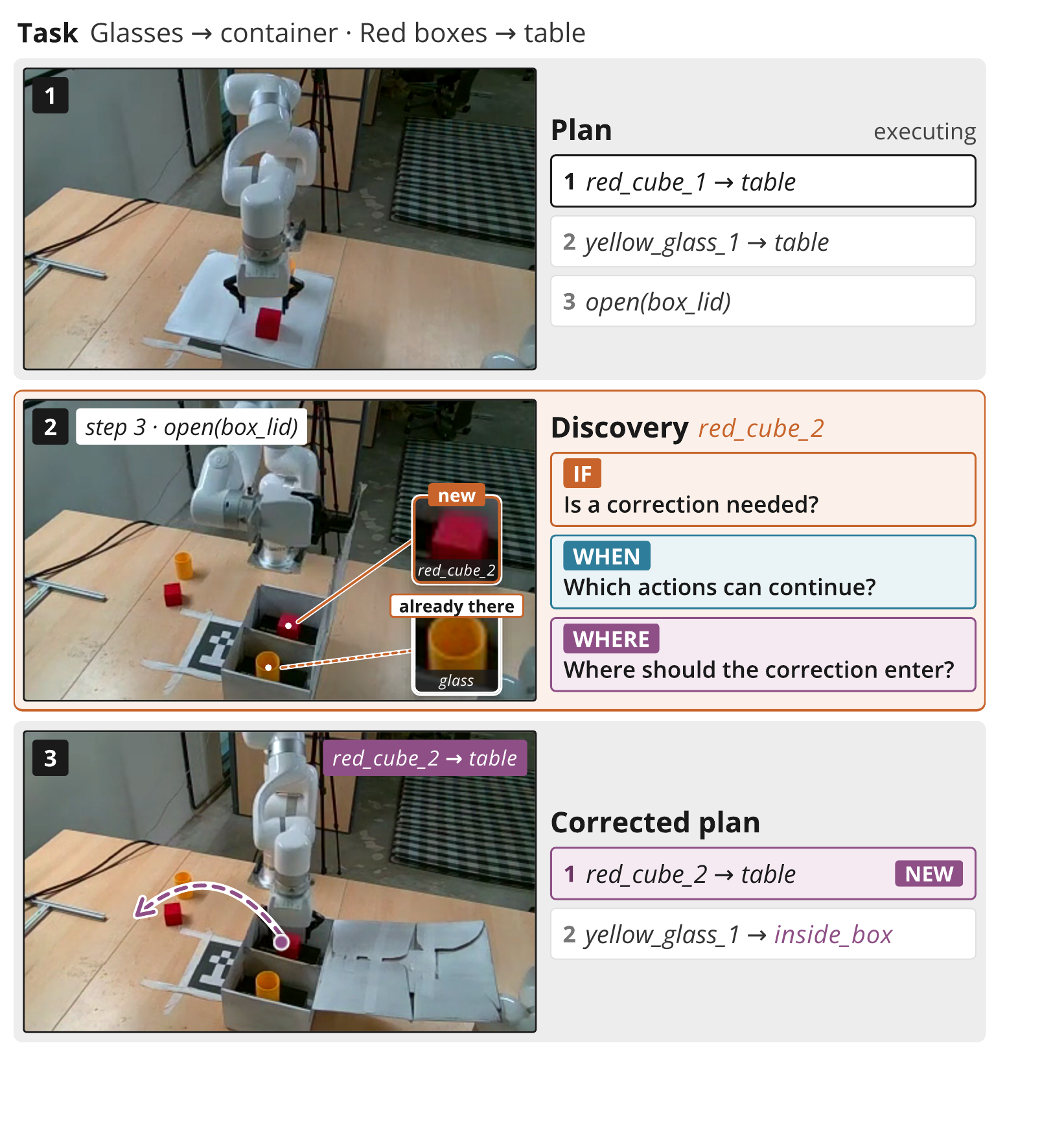}
    \caption{Opening a container reveals a hidden object, affecting the robot's ongoing plan and subsequent actions.}
    \label{fig:xARM-real}
\end{figure}
In long-horizon manipulation, task-relevant objects may not all be initially visible and may be revealed as the robot interacts with the scene. Consider a robot instructed to place all glasses (in yellow) inside the box and all blocks (in red) on the table (Figure ~\ref{fig:xARM-real}). Opening the container may reveal a hidden object that was absent from the initial observation and therefore omitted from the plan. The opening action succeeds, but completing the task now requires actions that the remaining plan does not cover.

Whether a discovery requires replanning depends on its relation to the task and the remaining actions. A glass mug already inside the container satisfies its placement requirement, whereas a task-irrelevant object like a phone occupying a planned placement area may need to be moved. An object may also be covered by actions that have not yet been executed. The robot must therefore reassess the remaining plan even though no action has failed.

When additional actions are needed, the correction must be coordinated with ongoing execution. In Figure~\ref{fig:xARM-real}, the newly discovered red block must be moved to the table, while the existing plan of placing the glass remains pending. The robot must therefore determine which actions can continue during replanning and where the additional transfer should enter the remaining action sequence. This requires incorporating newly discovered task requirements while preserving unaffected portions of the existing plan.

Preserving useful execution while revising a plan has long been studied in classical planning. A robot can refine upcoming steps using the state reached through execution~\cite{kaelbling2011hierarchical}, and repair a plan while continuing the portions that remain valid~\cite{lemai2004interleaving}. For manipulation, deciding whether an action remains executable also requires checking its physical feasibility. Task and motion planning (TAMP) addresses this requirement by considering action sequences together with the robot motions needed to execute them~\cite{garrett2020pddlstream}. These methods construct and revise plans using explicit descriptions of actions and their requirements. Foundation models (FMs) however, can infer intermediate goals from a natural-language instruction and the observed scene to guide task planning~\cite{yang2025guiding}. Closed-loop frameworks~\cite{huang2023inner} use changes in the scene and the success or failure of earlier actions to help the model decide what to do next. However, revising plans from feedback alone does not determine whether a newly discovered object requires correction, which actions can continue while that correction is planned, or where the corrective actions belong in the remaining sequence.

We introduce \textsc{Robust-TAMP}, a framework that decides \emph{if}, \emph{when}, and \emph{where} to replan as objects are discovered during manipulation. It first checks whether an observed object introduces an unmet task requirement or placement obstruction that the remaining plan does not address. When correction is needed, unaffected actions continue while the FM generates corrective action blocks. These blocks are validated and inserted according to their urgency, preserving the order of the remaining actions. We perform an extensive evaluation across 14 simulated task variants and five real-world scene variants, including comparisons with existing planners, component ablations, studies across FM size, input modality, reasoning mode, and prompting.
\section{Related work}
\label{sec:related}

\noindent \textbf{FMs for TAMP.} TAMP connects decisions about which actions to perform with the motions required to execute them. FMs support these decisions through different forms of scene input: LLM-based planners ~\cite{huang2023inner} use textual descriptions of observed objects and execution history, whereas VLM-based planners ~\cite{yan2025using} can also reason from images. Grounding language-based decisions in the robot's capabilities was demonstrated by SayCan~\cite{ichter2023saycan}, which combines language-model scores with skill value functions to select actions that are both useful and feasible. Planning over long-horizons also require information about the environment and the task completed thus far. LLM-Planner~\cite{song2023llmplanner} incorporates observed objects and completed subgoals into plan generation, while SayPlan~\cite{rana2023sayplan} uses hierarchical scene graphs to identify relevant parts of larger environments and simulator feedback to correct proposed plans. These methods use available robot skills, scene information, and feasibility checks to guide planning. Language instructions can also specify temporal constraints through AutoTAMP~\cite{chen2024autotamp} and goal, ordering, and blocking constraints through CaStL~\cite{guo2025castl}.

For manipulation, checking a proposed sequence also requires finding grasps, placements, and collision-free motions~\cite{wang2024llm3,curtis2025trust}. VLM-TAMP~\cite{yang2025guiding} connects visual reasoning to this process through intermediate goals, leaving the task and motion planner to find feasible actions that achieve them. When refinement fails, further model queries can revise those goals. OWL-TAMP~\cite{kumar2026owltamp} extends the model's role to proposing discrete plan structure and continuous constraints, allowing requirements inferred from language and images to guide motion planning. Together, these methods illustrate how FMs can specify what a planner should achieve and the conditions its solution should satisfy. 

\noindent \textbf{Planning with Incomplete Scene Knowledge.} Autonomous robots operating with incomplete scene knowledge must decide whether the available information is sufficient to proceed or further observations are needed. Belief-space methods account for uncertain states when selecting actions and revise plans as observations arrive ~\cite{hadfieldmenell2015modular,garrett2020online}. More directly concerned with hidden objects, planning under local observability~\cite{merlin2024robot} first seeks a solution using observed objects and initiates search when additional objects are needed to make the task feasible. This makes exploration depend on what the current scene description allows the robot to accomplish. EPoG~\cite{yang2026integrated} further integrates exploration with sequential manipulation, using observations and predicted object locations to update a scene graph and revise its plan. Its goal graph identifies the task-relevant objects in advance, even when their locations are unknown. Likewise, Azran et al.~\cite{azran2026bridging} connect visual perception
to belief-space planning by assigning probabilities to symbolic predicates, allowing the planner to account for uncertainty in the observed scene. These methods leave open how incidental object discoveries should change an ongoing plan, which we address by checking whether the remaining actions already account for the newly observed objects.

\noindent \textbf{Plan Repair and Concurrent Execution.} Replanning during execution must account for actions already underway and the time required to compute a replacement. Situated replanning~\cite{cashmore2019replanning} incorporates ongoing actions into the planning problem, including ways to interrupt them when necessary. Continuing execution during planning introduces a further trade-off: acting early can save time, but commits the robot before the complete plan is available. Concurrent planning has formalized this trade-off through metareasoning~\cite{elboher2023formal}, while temporal planning methods support dispatching actions before planning finishes under deadlines that affect action applicability and goal achievement~\cite{coles2024planning}. FM inference introduces a similar need to coordinate computation with execution. AgenticCache ~\cite{kim2026agenticcache} reduces repeated model calls by reusing cached plan transitions while an asynchronous process validates and updates them. These approaches address the timing and cost of planning, but do not directly determine where newly required actions should enter a plan already in progress. We address this by placing discovery-triggered corrections according to urgency and the execution progress made during their computation.
\section{Problem Formulation}
\label{sec:problem_formulation}
\begin{figure*}[!t]
    \centering
    \includegraphics[
        width=0.98\textwidth,
        trim={0cm 1cm 2.5cm 0cm},
        clip
    ]{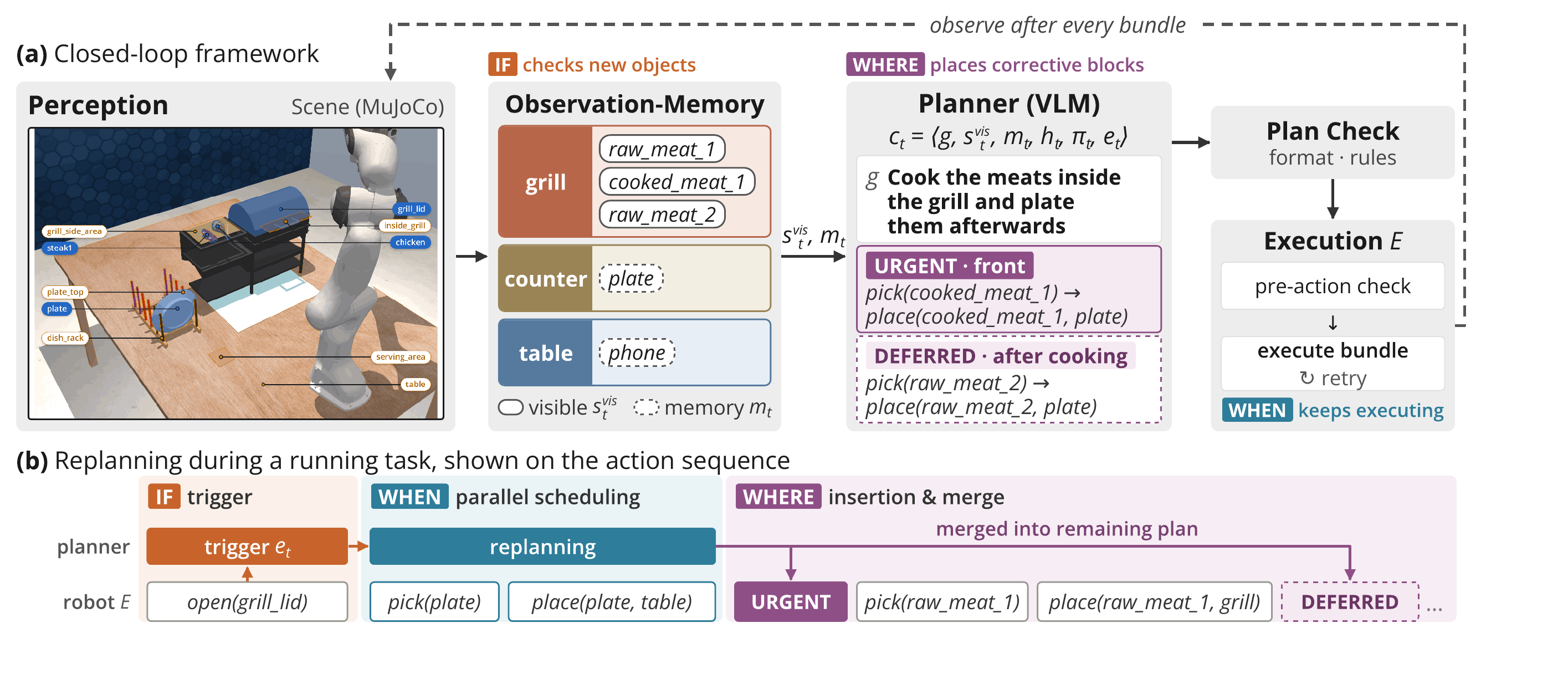}
    \vspace{-0.4cm}
    \caption{Overview of ROBUST-TAMP. Scene observations, memory, and execution history inform \textit{IF}: whether an observed object requires correction; \textit{WHEN}: which independent actions can continue
    during planning; and \textit{WHERE}: how corrective actions enter remaining plan.}
    \label{fig:iiith_overview}
\end{figure*}

We consider a robot given a natural-language goal $g$ to perform a manipulation task under incomplete knowledge of the objects in the scene. Specifically, we tackle scenarios where planned actions execute successfully,
but future observations may reveal previously unseen objects that are unaccounted for in the remaining plan. We consider two different domains, namely (1) the \textit{Kitchen domain} (Figure~\ref{fig:correction_examples}), which involves a container, and a cupboard, where the task is to place relevant grocery objects in the cupboard and mugs inside an initially closed container; and (2) the \textit{Grill domain} (Figure~\ref{fig:iiith_overview}), which require cooking raw meat and serving it on a plate. Further details of scene domain and variants are provided in Section~\ref{sec:experiments}. 
\subsection{Scene Knowledge and Observations}
Let $x_t$ denote the complete physical scene state at execution step
$t$, including the poses and states of all scene objects. The planner
does not have direct access to $x_t$. Instead, it receives an observed
scene description $s_t^{\mathrm{vis}}$ containing the currently visible
objects, their object--region relations, relevant articulation states,
and placement regions. Memory $m_t$ retains information from the current and previous
observations. We write its update compactly as
\begin{equation}
    m_t =
    \textsc{UpdateMemory}
    \left(m_{t-1},s_t^{\mathrm{vis}}\right),
    \label{eq:memory}
\end{equation}
with $m_{-1}=\emptyset$. For each observed object, memory records its
most recently observed region and observation time. An object is
newly discovered when it appears in the current observation but has
no previous entry in $m_{t-1}$. Thus, an object that merely re-enters
view is not treated as a new discovery. The history $h_t$ records completed actions in execution order.
\subsection{Revising an Executing Plan}
Let $\pi$ be the initial plan synthesized. We define the remaining plan $\pi_t$ as the ordered sequence of actions that have not yet been executed. As execution progresses, completed actions
are removed from $\pi_t$. When a discovery requires correction, new
actions may be inserted while preserving the unaffected portions of
the remaining plan. Actions specify supported manipulation operations
and, when applicable, their objects and destination regions. The
execution backend $E$ determines the corresponding grasps, placements,
and feasible motions.
\begin{table}[t]
    \centering
    \caption{Core notation used by \textsc{ROBUST-TAMP}.}
    \label{tab:notation}
    \small
    \begin{tabular}{@{}ll@{}}
        \toprule
        Symbol & Meaning \\
        \midrule
        $g$ & Natural-language task goal \\
        $s_t^{\mathrm{vis}}$ & Current observed scene description \\
        $m_t$ & Memory of previous observations \\
        $h_t$ & Completed-action history \\
        $\pi_t$ & Current remaining action sequence \\
        $c_t$ & Planning context supplied to the FM \\
        $e_t$ & Current planning event \\
        $r_t$ & FM planning response \\
        $E$ & Execution backend \\
        \bottomrule
    \end{tabular}
\end{table}

A discovery may introduce an unmet task requirement, obstruct a
pending action, or require no correction. Given
$(g,s_t^{\mathrm{vis}},m_t,h_t,\pi_t)$, the problem is to determine
whether revision is necessary and, when it is, produce an executable
continuation. This requires deciding which pending actions can proceed
while a correction is being computed and where the corrective actions
should enter the remaining sequence. Because execution may continue during planning, the correction must be merged with the current $\pi_t$ after accounting for actions that completed during its computation.
\subsection{Goal Satisfaction}
Let $\tau$ denote a task instance, comprising the goal $g$
and the initial scene. We denote by $\mathcal{R}_{\tau}$ the required final object--region relations and by $\mathcal{P}_{\tau}$ the procedural conditions evaluated over the execution history. Success of $\tau$ at termination $T$ is then determined by
\begin{equation}
    G_{\tau}(x_T,h_T)=
    \bigwedge_{(o,r)\in\mathcal{R}_{\tau}}
    [\operatorname{region}(x_T,o)=r]
    \land
    \bigwedge_{p\in\mathcal{P}_{\tau}}p(h_T).
    \label{goal}
\end{equation}
The first term requires every specified object to occupy its target region; the second requires the execution history to satisfy every prescribed procedure. For example, in the grill domain, placing meat on the serving plate is insufficient unless the execution history also records that it was cooked beforehand. Both conditions must hold for task success, distinguishing correct final arrangements from tasks that additionally require a particular sequence of operations. Success is evaluated using the complete physical state and execution history. Since the planner has only partial scene
knowledge, its current plan may omit requirements  involving unobserved objects--- executing this plan does not necessarily complete the task.  

\section{Methodology}
\label{sec:methodology}
\textsc{ROBUST-TAMP} makes three decisions as new
scene information becomes available during execution: \textit{IF} an
observed object warrants a plan correction, \textit{WHEN} unaffected actions
can continue while that correction is computed, and \textit{WHERE} the
corrective actions should enter the remaining plan. Figure~\ref{fig:iiith_overview}(a) shows the closed-loop framework, while Figure~\ref{fig:iiith_overview}(b) illustrates these decisions during an executing task. Algorithm~\ref{alg:robust_tamp} summarizes
the execution loop, and Table~\ref{tab:notation} defines the notation used throughout our work.




\begin{algorithm}[t]
\caption{\textsc{ROBUST-TAMP}}
\label{alg:robust_tamp}
\begin{algorithmic}[1]
\footnotesize
\Require Goal $g$, executor $E$, FM-call budget

\State $s^{\mathrm{vis}},m,h,\pi \gets \emptyset$;
       $e \gets \textsc{Initial}$
       \label{alg:init}

\State $(s^{\mathrm{vis}},m)
       \gets \textsc{ObserveUpdate}(s^{\mathrm{vis}},m)$
       \label{alg:observe}

\While{$e$ is not terminal and budget remains}

    \State $c \gets
    \langle g,s^{\mathrm{vis}},m,h,\pi,e\rangle$
    \label{alg:context}

    \If{$e$ is a discovery trigger}

        \State $(r,e) \gets
        \textsc{PlanWhileExecuting}(c,E)$
        \Comment{\textsc{WHEN}: non-interfering}
        \label{alg:when}

        \If{$e$ is an execution failure}
            \label{alg:when-fail-cond}
            \State \textbf{continue}
            \label{alg:when-fail}
        \EndIf

    \Else
        \State $r \gets \textsc{QueryFM}(c)$
    \EndIf

    \State $(\pi,e) \gets
    \textsc{ValidateAndMerge}(r,c)$
    \Comment{\textsc{WHERE}: urgency / insertion}
    \label{alg:where}

    \If{$e\neq\emptyset$}
        \State \textbf{continue}
        \label{alg:validation-fail}
    \EndIf

    \While{$\pi\neq\langle\rangle$ and $e=\emptyset$}

        \State $e \gets \textsc{ExecuteMonitor}(E)$
        \Comment{\textsc{IF}: unmet / obstructing}
        \label{alg:if}

    \EndWhile

    \If{$\pi=\langle\rangle$ and $e=\emptyset$}
        \State \Return $h$
        \label{alg:exhausted}
    \EndIf

\EndWhile

\State \Return $h$

\end{algorithmic}
\end{algorithm}

\subsection{Planning Interface}

At execution step $t$, the framework maintains the current observed scene $s_t^{\mathrm{vis}}$, memory $m_t$, completed-action history $h_t$, and remaining plan $\pi_t$. The goal $g$ remains fixed throughout the task. \textsc{ObserveUpdate} (Algorithm~\ref{alg:robust_tamp}, line~\ref{alg:observe}) refreshes the current scene observation and updates memory according to~\eqref{eq:memory}. When an object leaves view, its most recently observed region and observation time remain in $m_t$, allowing previously observed information to be distinguished from current visual evidence. In simulation, segmentation determines visibility, while simulator
geometry supplies region assignments and placement-area overlap for observed objects.

Planning is initiated by an event $e_t$. The corresponding planning context is
\[
c_t =
\left\langle
g,\,
s_t^{\mathrm{vis}},\,
m_t,\,
h_t,\,
\pi_t,\,
e_t
\right\rangle ,
\]
as constructed in Algorithm~\ref{alg:robust_tamp},
line~\ref{alg:context}, and illustrated in Figure~\ref{fig:iiith_overview}(a). The event $e_t$ specifies the reason for invoking the planner: initial planning, discovery-triggered correction, validation feedback, or recovery from an execution failure. The FM response $r_t$ contains either a complete
action sequence for initial planning or failure recovery, or a set of
corrective action blocks for a discovery-triggered request. If no
corrective action is required, the FM may instead return
\mbox{\textnormal{\textsc{NO\_ACTIONS}}}. LLM and VLM planners receive
the same structured task information, while VLMs additionally receive
camera observations.

Initial planning produces an ordered sequence of actions: \texttt{pick}, \texttt{place}, and \texttt{open} for \textit{kitchen} tasks, with \texttt{close} additionally supported in \textit{grill} tasks to complete cooking procedure. Before execution, the returned sequence is parsed to check action names, object and region references, and manipulation-sequence consistency. The execution backend $E$ then determines grasps, placements, and feasible
motions and verifies action preconditions and outcomes. Execution proceeds in bundles, with adjacent pick and place actions for the same object treated as one bundle. Completed actions are appended to
$h_t$ and removed from $\pi_t$, after which the scene and memory are updated before execution proceeds. The procedures in Algorithm~\ref{alg:robust_tamp} update $s^{\mathrm{vis}}$, $m$, $h$, and $\pi$ in place; the algorithm omits the subscript $t$ when referring to their current values. Termination of the execution loop does not necessarily indicate task success; the returned execution history and final scene state are evaluated against the goal conditions in~\eqref{goal}.
\subsection{\textit{IF}: Selective Replanning}
The \textit{IF} decision checks whether any observed object requires additional actions that are not already included in the remaining plan. After each completed execution bundle, \textsc{ExecuteMonitor} (Algorithm~\ref{alg:robust_tamp}, line~\ref{alg:if}) reassesses the currently observed objects together with objects retained in memory. A newly visible object therefore does not automatically trigger replanning; the framework first determines whether any additional action is required.

The first check is whether the remaining plan already accounts for the object's task requirements. An object is considered covered when its pending actions in $\pi_t$ address its required placement or procedure, rather than merely involving the object. Such objects do not trigger additional actions, as illustrated in Figure~\ref{fig:iiith_overview}(b). Similarly, a previous \texttt{NO\_ACTIONS} decision remains valid while the object's recorded region and its relation to the remaining plan are unchanged. The object is reconsidered if its region changes or subsequent plan revisions introduce new requirements or placement conflicts. These checks avoid repeated FM calls while allowing earlier decisions to be revisited when their underlying conditions change.

For each uncovered object, the framework evaluates its relation to the task and the remaining plan. A task-relevant object requires correction when its task requirement is still unmet, regardless of whether it affects another action. For example, a grocery discovered outside the cupboard introduces additional actions because its required final relation has not yet been achieved. In contrast, a task-irrelevant object requires correction only when it occupies a placement area required by a pending action. Thus, a phone occupying a mug-placement area must be moved, whereas the same phone elsewhere causes no change to the plan. Determining whether a task-relevant object has satisfied its goal requirements may also depend on the completed-action history $h_t$. In the grill domain, for example, placing meat on the serving plate satisfies the goal only if the required cooking procedure has already been completed. For objects that are no longer visible, the framework uses their last known locations stored in $m_t$ to determine whether they still require additional actions.

Consequently, a correction is requested only when an object is not
already covered and either (i) it is task-relevant with an unmet
requirement, or (ii) it is task-irrelevant but blocks a placement
required by the remaining plan. All such objects are
grouped into a single discovery-triggered event $e_t$. If no object
satisfies the rule, execution continues without an additional FM call.

\begin{figure*}[!t]
    \centering
    \includegraphics[
        width=0.98\textwidth,
        trim={0cm 0.8cm 2.25cm 0cm},
        clip
    ]{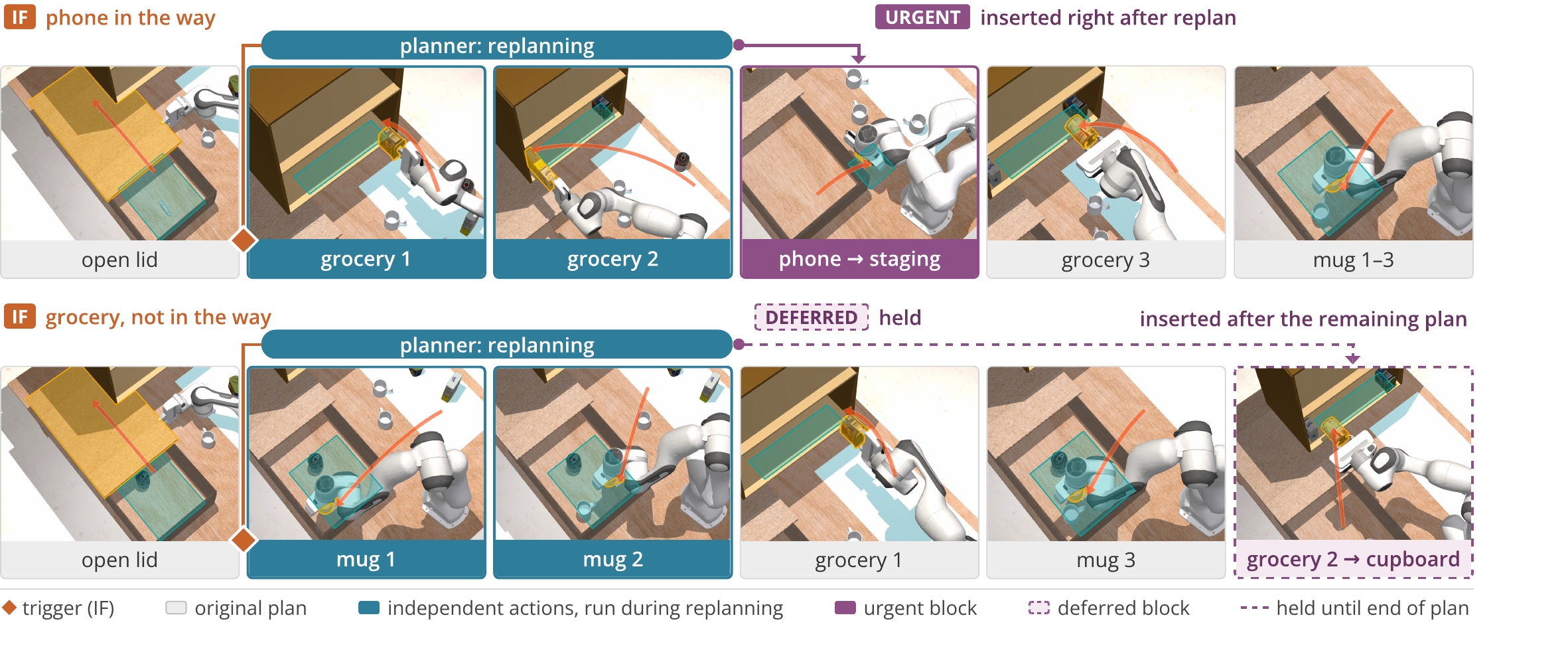}
    \vspace{-0.4cm}
    \caption{Urgent and deferred corrections during execution. Independent actions continue while FM plans a correction. Urgent blocks enter before the updated remaining plan; deferred blocks follow selected pending action or are appended to the plan.}
    \label{fig:correction_examples}
\end{figure*}
\subsection{\textit{WHEN}: Concurrent Planning and Execution}

Once \textit{IF} raises a discovery-triggered event,
\textsc{PlanWhileExecuting}
(Algorithm~\ref{alg:robust_tamp}, line~\ref{alg:when})
requests corrective actions from the FM while allowing unaffected
parts of the remaining plan to continue. The \textit{WHEN} decision
therefore determines which pending execution bundles can safely
proceed before the corrective response is available. The scheduler first identifies the objects and regions that the correction may involve, including the triggering objects, their task-goal regions, obstructed placement areas, and designated staging regions. Each remaining bundle is then checked using its manipulated object, source region, and destination region. A bundle is withheld when executing it could conflict with an object or region reserved for the correction. The scheduler scans $\pi_t$ in execution order. When a bundle is
withheld, later bundles that depend on its objects or regions are also
held back, while independent bundles continue in their original
relative order. For example, if a newly observed phone blocks the
container used for mug placement, those mug actions can wait while
independent grocery transfers to the cupboard continue, as illustrated
in Figure~\ref{fig:iiith_overview}(b) and Figure~\ref{fig:correction_examples}.

Bundles selected for concurrent execution are listed separately in the
FM request. Their anticipated effects are not incorporated into
$s_t^{\mathrm{vis}}$ before they actually occur, preventing the planner
from reasoning over state changes that have not yet been executed.
Observations and execution checks continue while the FM is running.
When the response becomes available, the currently executing bundle is
allowed to finish before the correction is processed; if all eligible
bundles finish first, the robot waits for the response. If a concurrently executed bundle fails, the corrective response is discarded (Algorithm~\ref{alg:robust_tamp}, lines~\ref{alg:when-fail-cond}--\ref{alg:when-fail}), and the execution failure becomes the next planning event.

\subsection{\textit{WHERE}: Urgency-Based Insertion}

The \textit{WHERE} decision determines where newly required actions
enter the remaining plan. For a discovery-triggered request, the
FM response $r_t$ contains one or more corrective blocks, each
specifying the objects addressed, the required actions, an urgency
label, and a proposed insertion position as illustrated in Figure~\ref{fig:correction_examples}.

Urgency is determined by what would happen if the correction were
delayed. A block is urgent when postponing it would prevent a pending
action from executing correctly or violate a task procedure.
Such a block is placed at the front of the remaining plan and executes
at the next available bundle boundary. If delaying the correction
does not invalidate the intervening work, the block is deferred and
inserted after an eligible pending action or at the end of the plan.
In either case, the existing pending actions otherwise retain their
relative order. Discovery-triggered replanning therefore augments the
current plan without regenerating unaffected portions.

Before accepting the proposed correction,
\textsc{ValidateAndMerge} (Algorithm~\ref{alg:robust_tamp}, line~\ref{alg:where}) checks that every triggering object is addressed, all actions, objects, and regions are supported, and each corrective block forms a complete manipulation sequence. If a correction is deferred, its insertion point is chosen from the actions that are still waiting to execute after concurrent execution. Actions selected to execute during FM inference are not considered as insertion points. This prevents corrective actions from being inserted after they are needed. Procedural urgency that cannot be determined from such geometric conflicts, such as removing already cooked meat before a subsequent cooking cycle, remains a decision made by the FM rather than being
recomputed by the validator.

Actions completed during FM inference are first removed from the remaining plan, after which the accepted corrective blocks are merged into the resulting $\pi_t$. Monitored execution then resumes from this updated sequence. If validation fails, the feedback becomes the next
planning event $e_t$
(Algorithm~\ref{alg:robust_tamp},
lines~\ref{alg:where}--\ref{alg:validation-fail}),
and the FM is queried again while the current remaining plan is
preserved.

\section{Experiments and Evaluations}
\label{sec:experiments}
\subsection{Experimental Setup}
All FM inference runs on the same NVIDIA RTX 5090 server with 32\,GB GPU memory through vLLM.
For the selected planner (Qwen3-VL-8B-Thinking; see \hyperref[RQ1]{\textbf{RQ1}}), temperature is 1.0, top-$p$ is 0.95, top-$k$ is 20, and the output limit is 24,576 tokens. ROBUST TAMP permits up to ten
replanning calls after the initial planning request. The planning interface, observation memory, and \textit{IF}, \textit{WHEN}, and \textit{WHERE} mechanisms are shared across simulation and real-world execution. The supplementary material provides the source code, planning prompts, in-context examples, baseline adaptation details, model and inference configurations, and \href{https://narendhiranv04.github.io/ROBUST-TAMP/}
{execution videos}\footnote{\url{https://narendhiranv04.github.io/ROBUST-TAMP/}}.

\noindent{\textbf{Simulation Scenes and Variants.}}
We evaluate ROBUST TAMP on 14 MuJoCo task variants across kitchen and grill domains (Table~\ref{tab:main}). The kitchen task requires placing groceries in a cupboard and mugs inside a lidded container. The grill task requires cooking raw meat and serving cooked meat on a plate in the serving area. The variants cover different reasons for revising an ongoing plan. Basic variants K0 and G0 contain no hidden objects. Kitchen variants K1--K4 combine task relevance and placement-area overlap, covering obstruction removal, discoveries requiring no correction, and urgent or deferred handling of additional groceries. Grill variants G1--G3 reveal cooked meat, mixed cooked and raw meat, or raw meat, respectively, requiring corrections that respect cooking and serving order. Cardinality variants increase the number of hidden groceries (K3-n2, K3-n3), reduce the number of hidden cooked meats (G1-n1), or add independent groceries that can be handled during replanning (K1-w1, K1-w2).

Simulator segmentation determines object visibility, while simulator geometry provides region assignments and placement-area overlap for observed objects.
Hidden-object information is withheld from the planner until the objects become visible. In simulation, a Franka Panda arm executes action primitives using damped-least-squares inverse kinematics and RRT-Connect~\cite{kuffner2000rrtconnect} motion planning. We use ten trials per variant with seeds 0--9, giving 140 trials across the full suite. Component ablations use the seven core variants,
K1--K4 and G1--G3.
\begin{table*}[t]
\centering
\begin{threeparttable}
\caption{Planner comparison across (a) model size, modality, and prompting, and (b) reasoning and non-reasoning 8B models.}
\label{tab:planner-selection}
\scriptsize
\setlength{\tabcolsep}{1.6pt}
\renewcommand{\arraystretch}{1.03}
\begin{tabular*}{\textwidth}{
@{\extracolsep{\fill}}
lll ccc ccc r
@{\hspace{8pt}}
lcl ccc ccc r
@{}
}
\toprule
\multicolumn{10}{c}{(a) Scale, modality, and prompting}
&
\multicolumn{10}{c}{(b) 8B-class models}
\\
\cmidrule(r){1-10}
\cmidrule(l){11-20}

& & &
\multicolumn{3}{c}{SR}
&
\multicolumn{3}{c}{PGC}
& &
& & &
\multicolumn{3}{c}{SR}
&
\multicolumn{3}{c}{PGC}
&
\\
\cmidrule(lr){4-6}
\cmidrule(lr){7-9}
\cmidrule(lr){14-16}
\cmidrule(lr){17-19}

Scale & Mod. & Prompt
& K & G & All
& K & G & All
& Plan.
&
Mod. & Reas. & Model
& K & G & All
& K & G & All
& Plan.
\\
\midrule

\multirow{4}{*}{4B}
& \multirow{2}{*}{LLM}
& ZS
& 14.4 & 0.0 & 9.3
& 18.0 & 41.5 & 26.4
& 242
&
\multirow{6}{*}{VLM}
& \multirow{3}{*}{--}
& Qwen3-VL-8B-Instruct
& 1.1 & 2.0 & 1.4
& 2.9 & 24.5 & 10.6
& 168
\\

& & ICL
& \textcolor{gray}{14.4} & 14.0 & 14.3
& \textcolor{gray}{18.0} & 54.5 & 31.1
& 206
&
& & Ministral-3-8B-Instruct
& 16.7 & 2.0 & 11.4
& 54.2 & 52.7 & 53.7
& 47
\\
\cmidrule(r){2-10}

& \multirow{2}{*}{VLM}
& ZS
& 53.3 & 6.0 & 36.4
& 55.5 & 43.8 & 51.3
& 736
&
& & InternVL3.5-8B
& 18.9 & 0.0 & 12.1
& 28.1 & 20.1 & 25.2
& 111
\\
\cmidrule(l){12-20}

& & ICL
& \textcolor{gray}{53.3} & 70.0 & 59.3
& \textcolor{gray}{55.5} & 90.1 & 67.9
& 560
&
& \multirow{3}{*}{$\checkmark$}
& \textbf{Qwen3-VL-8B-Thinking}
& \textbf{94.4} & \textbf{22.0} & \textbf{68.6}
& \textbf{97.7} & \textbf{66.3} & \textbf{86.5}
& \textbf{540}
\\
\cmidrule(r){1-10}

\multirow{4}{*}{8B}
& \multirow{2}{*}{LLM}
& ZS
& 81.1 & 4.0 & 53.6
& 83.9 & 40.3 & 68.3
& 327
&
& & Ministral-3-8B-Reasoning
& 48.9 & 70.0 & 56.4
& 51.9 & 89.1 & 65.1
& 468
\\

& & ICL
& \textcolor{gray}{81.1} & 34.0 & 64.3
& \textcolor{gray}{83.9} & 58.2 & 74.7
& 257
&
& & Holo2-8B
& 31.1 & 2.0 & 20.7
& 41.7 & 37.7 & 40.2
& 135
\\
\cmidrule(r){2-10}
\cmidrule(l){11-20}

& \multirow{2}{*}{VLM}
& ZS
& 94.4 & 22.0 & 68.6
& 97.7 & 66.3 & 86.5
& 540
&
\multirow{6}{*}{LLM}
& \multirow{3}{*}{--}
& Qwen3-8B (no think)
& 1.1 & 6.0 & 2.9
& 4.6 & 28.3 & 13.1
& 25
\\

& & ICL
& \textcolor{gray}{\textbf{94.4}}
& \textbf{88.0} & \textbf{92.1}
& \textcolor{gray}{\textbf{97.7}}
& \textbf{98.1} & \textbf{97.8}
& \textbf{355}
&
& & Llama-3.1-8B-Instruct
& 6.7 & 0.0 & 4.3
& 12.2 & 18.7 & 14.5
& 124
\\
\cmidrule(r){1-10}

\multirow{4}{*}{32B}
& \multirow{2}{*}{LLM}
& ZS
& 91.1 & 18.0 & 65.0
& 93.0 & 69.4 & 84.6
& 390
&
& & Qwen2.5-7B-Instruct
& 0.0 & 0.0 & 0.0
& 2.8 & 12.6 & 6.3
& 10
\\
\cmidrule(l){12-20}

& & ICL
& \textcolor{gray}{91.1} & 66.0 & 82.1
& \textcolor{gray}{93.0} & 92.9 & 93.0
& 185
&
& \multirow{3}{*}{$\checkmark$}
& Qwen3-8B (think)
& 81.1 & 4.0 & 53.6
& 83.9 & 40.3 & 68.3
& 327
\\
\cmidrule(r){2-10}

& \multirow{2}{*}{VLM}
& ZS
& 91.1 & 30.0 & 69.3
& 93.9 & 76.4 & 87.7
& 1107
&
& & R1-Distill-Llama-8B
& 0.0 & 0.0 & 0.0
& 1.9 & 12.6 & 5.7
& 500
\\

& & ICL
& \textcolor{gray}{91.1} & 82.0 & 87.9
& \textcolor{gray}{93.9} & 96.0 & 94.7
& 603
&
& & R1-Distill-Qwen-7B
& 0.0 & 0.0 & 0.0
& 2.5 & 20.1 & 8.8
& 597
\\
\bottomrule
\end{tabular*}

\begin{tablenotes}[flushleft]
\scriptsize
\item \textbf{SR/PGC}: success/partial goal completion (\%);
\textbf{K/G}: kitchen/grill; \textbf{All}: all 14 variants; \textbf{Plan}.: mean planning time per trial (s). \textbf{Bold} marks the selected configuration in each panel. \textbf{ZS}: zero-shot; \textbf{ICL}: examples supplied only to grill prompts. Grey kitchen values are shared with ZS. \textbf{Reas.}: reasoning enabled ($\checkmark$) or disabled (--).
\end{tablenotes}

\end{threeparttable}
\end{table*}

\noindent{\textbf{Real-World Implementation.}} The physical implementation (Figure~\ref{fig:xARM-real}) uses a UFACTORY X-ARM 7 manipulator and two overhead D545 cameras. SAM 3~\cite{carion2025sam3} provides object segmentation masks, and observed objects are associated with predefined placement regions. These object--region observations update the same memory layer, retaining the last observed information when an object leaves view. Point clouds obtained from the cameras support grasp generation, after which the resulting manipulation actions are executed on the robot. We examine five variants: nominal execution, task-irrelevant discoveries with and without placement conflicts, and task-relevant discoveries requiring urgent or deferred corrections.
\subsection{Evaluation Metrics}
For trial $i$, let $c_i$ be the number of required final relations and cooking conditions, and $u_i$ the number unsatisfied at termination. Over $N$ trials, success rate (SR) and partial goal completion (PGC) are computed as
\begin{equation}
\mathrm{SR}
= \frac{100}{N}\sum_{i=1}^{N}\mathbf{1}[u_i=0],
\qquad
\mathrm{PGC}
= \frac{100}{N}\sum_{i=1}^{N}
  \left(1-\frac{u_i}{c_i}\right).
\label{eq:outcome_metrics}
\end{equation}
SR$_K$ and SR$_G$ report kitchen and grill success separately. Outcomes are evaluated from the final scene and execution history; exhausting a plan does not itself establish success. Planning and execution errors are analyzed separately in \hyperref[RQ1]{\textbf{RQ4}}.

Let $n_i$ denote FM calls, $\ell_{ij}$ the duration of call $j$, and $T_i$ the total trial duration. For trial $i$, let $r_i$ be the number of discovery-triggered
replanning episodes. In episode $k$, $w_{ik}$ measures the time the robot waits for a corrective response, including validation retries and excluding time during which execution continues. Thus, Idle reports mean waiting time per discovery-triggered episode; it excludes initial planning and replanning prompted by execution failures. Times are in seconds, and trial averages include unsuccessful runs.
\begin{equation}
\setlength{\jot}{5pt}
\begin{gathered}
\mathrm{Calls}=\frac{\sum_i n_i}{N},\quad
\mathrm{Plan.}=\frac{\sum_i\sum_{j=1}^{n_i}\ell_{ij}}{N},\quad
\mathrm{Time}=\frac{\sum_i T_i}{N},\qquad
\\
\mathrm{Lat.}=\operatorname{median}_{i,j}\ell_{ij}, \quad
\mathrm{Idle}=\frac{\sum_i\sum_{k=1}^{r_i}w_{ik}}{\sum_i r_i}.
\end{gathered}
\label{eq:cost_metrics}
\end{equation}

\begin{table*}[t]
\centering
\begin{threeparttable}
\caption{Task variants and \textsc{ROBUST-TAMP} performance across
basic, core, and cardinality settings.}
\label{tab:main}

\footnotesize
\setlength{\tabcolsep}{2pt}
\renewcommand{\arraystretch}{1.0}

\begin{tabular*}{\textwidth}{
@{\extracolsep{\fill}}
ll ccc c ccccccc
@{}
}
\toprule
ID & Hidden
& Rel. & Ovl. & Exp.
& Ceil.
& SR & PGC & Calls & Plan. & Idle & Time & Urg.
\\
\midrule

\multicolumn{13}{@{}l}{\textit{A. Basic}}
\\
K0 & None
& -- & -- & --
& 100.0
& 100.0 & 100.0 & 2.2 & 229 & -- & 298 & --
\\
G0 & None
& -- & -- & --
& 100.0
& 100.0 & 100.0 & 1.1 & 117 & -- & 197 & --
\\
\midrule

\multicolumn{13}{@{}l}{\textit{B. Core}}
\\
K1 & Phone
& $\times$ & $\checkmark$ & U
& 100.0
& 90.0 & 93.3 & 5.7 & 481 & 189 & 561 & 10.0
\\
K2 & Phone
& $\times$ & $\times$ & --
& 100.0
& 90.0 & 95.0 & 4.0 & 328 & -- & 397 & --
\\
K3 & Grocery
& $\checkmark$ & $\checkmark$ & U
& 100.0
& 100.0 & 100.0 & 3.8 & 385 & 122 & 480 & 70.0
\\
K4 & Grocery
& $\checkmark$ & $\times$ & D
& 100.0
& 100.0 & 100.0 & 3.1 & 316 & 96 & 390 & 10.0
\\
G1 & 2 cooked meat
& $\checkmark$ & $\times$ & U
& 100.0
& 80.0 & 97.1 & 2.2 & 204 & 97 & 296 & 90.0
\\
G2 & Cooked + raw meat
& $\checkmark$ & $\times$ & U/D
& 100.0
& 90.0 & 98.6 & 2.5 & 285 & 139 & 377 & 90.0
\\
G3 & 2 raw meat
& $\checkmark$ & $\times$ & D
& 100.0
& 90.0 & 98.6 & 3.4 & 355 & 169 & 445 & 50.0
\\
\midrule

\multicolumn{13}{@{}l}{\textit{C. Cardinality}}
\\
K3-n2 & 2 groceries
& $\checkmark$ & $\checkmark$ & U
& 100.0
& 90.0 & 98.6 & 3.9 & 413 & 143 & 522 & 70.0
\\
K3-n3 & 3 groceries
& $\checkmark$ & $\checkmark$ & U
& 100.0
& 100.0 & 100.0 & 4.0 & 442 & 128 & 580 & 86.7
\\
G1-n1 & 1 cooked meat
& $\checkmark$ & $\times$ & U
& 100.0
& 80.0 & 96.0 & 2.6 & 248 & 132 & 328 & 50.0
\\
K1-w1 & Phone; +1 grocery
& $\times$ & $\checkmark$ & U
& 100.0
& 90.0 & 98.6 & 7.7 & 655 & 266 & 774 & 0.0
\\
K1-w2 & Phone; +2 groceries
& $\times$ & $\checkmark$ & U
& 100.0
& 90.0 & 93.8 & 6.4 & 515 & 143 & 648 & 0.0
\\
\midrule

\multicolumn{5}{@{}l}{Mean (kitchen)}
& 100.0
& 94.4 & 97.7 & 4.5 & 418 & 144 & 517 & 49.0
\\
\multicolumn{5}{@{}l}{Mean (grill)}
& 100.0
& 88.0 & 98.1 & 2.4 & 242 & 134 & 329 & 72.9
\\
\multicolumn{5}{@{}l}{\textbf{Mean (all)}}
& \textbf{100.0}
& \textbf{92.1}
& \textbf{97.8}
& \textbf{3.8}
& \textbf{355}
& \textbf{141}
& \textbf{449}
& \textbf{58.8}
\\
\bottomrule
\end{tabular*}

\begin{tablenotes}[flushleft]
\scriptsize
\item \textbf{+$w$} denotes additional independent groceries initially on the table. \textbf{Rel.}: task relevance; \textbf{Ovl}.: placement-area overlap; \textbf{Exp.}: expected correction (\textbf{U}: urgent, \textbf{D}: deferred, \textbf{--}: no discovery-triggered correction). \textbf{Ceil.}: observed oracle-planner success with the same executor. \textbf{Urg.}: percentage of evaluated trigger objects assigned the correct urgency on the first attempt.
\end{tablenotes}

\end{threeparttable}
\end{table*}

\begin{table}[t]
\centering
\begin{threeparttable}

\caption{Ablation study of our proposed \textsc{ROBUST-TAMP}.}
\label{tab:ablations}

\footnotesize
\renewcommand{\arraystretch}{1.05}
\setlength{\tabcolsep}{1.2pt}

\begin{tabular*}{\columnwidth}{
@{\extracolsep{\fill}}
l
ccccc
ccccc
@{}
}
\toprule
&
\multicolumn{5}{c}{\textit{Kitchen}}
&
\multicolumn{5}{c}{\textit{Grill}}
\\
\cmidrule(lr){2-6}
\cmidrule(lr){7-11}

Condition
& SR & C & I & P & T
& SR & C & I & P & T
\\
\midrule

\textbf{Full}
& \textbf{95.0} & 4.2 & 118 & 377 & 457
& \textbf{86.7} & 2.7 & 135 & 281 & 373
\\
\midrule

$-$Mem.
& 85.0 & 4.0 & 131 & 342 & 420
& 60.0 & 2.6 & 175 & 343 & 435
\\

$-$IF
& 95.0 & 5.5 & 199 & 441 & 511
& 76.7 & 2.8 & 171 & 317 & 408
\\

$-$WHERE
& 75.0 & 3.9 & 76 & 348 & 434
& 56.7 & 4.5 & 221 & 546 & 648
\\

Front
& 90.0 & 4.0 & 116 & 394 & 473
& 30.0 & 6.1 & 185 & 821 & 916
\\

End
& 52.5 & 6.5 & 438 & 562 & 609
& 30.0 & 2.5 & 188 & 320 & 410
\\

$-$WHEN
& 92.5 & 4.5 & 127 & 399 & 483
& 66.7 & 2.5 & 147 & 274 & 375
\\

LLM
& 87.5 & 5.2 & 65 & 205 & 286
& 23.3 & 9.2 & 288 & 375 & 426
\\

\bottomrule
\end{tabular*}

\begin{tablenotes}[flushleft]
\scriptsize
\item
Kitchen: K1--K4, zero-shot; Grill: G1--G3, ICL.
\textbf{C}: FM calls;
\textbf{I}: idle time;
\textbf{P}: planning time;
\textbf{T}: total trial time.
Times are in seconds.
\textbf{$-$Mem.}: current observations only;
\textbf{$-$IF}: replan for every newly visible object;
\textbf{$-$WHERE}: regenerate the remaining plan;
\textbf{Front/End}: fixed corrective insertion;
\textbf{$-$WHEN}: suspend execution during replanning;
\textbf{LLM}: Qwen3-8B planner.
\end{tablenotes}

\end{threeparttable}
\end{table}

\subsection{Results and Discussion}
\newcommand{\icl}{\hspace{0.6em}\textit{+\,ICL}}
\begin{table}[h]
\centering
\footnotesize
\begin{threeparttable}
\setlength{\tabcolsep}{2.2pt}
\renewcommand{\arraystretch}{1.05}
\caption{Baseline Comparison across 14 MuJoCo variants}
\label{tab:baselines}
\begin{tabular*}{\columnwidth}{@{\extracolsep{\fill}}l cccc cccc@{}}
\toprule
& \multicolumn{4}{c}{Success (\%) $\uparrow$} & \multicolumn{4}{c}{Cost $\downarrow$} \\
\cmidrule(lr){2-5}\cmidrule(l){6-9}
Method & SR$_K$ & SR$_G$ & SR & PGC & Calls & Plan. & Lat. & Time \\
\midrule
VLM-TAMP~\cite{yang2025guiding}               & \multirow{2}{*}{23.3} & 6.0 & 17.1 & 71.4 & 2.1 & 77 & 31.8 & 176 \\
\icl                                   & & 10.0 & 18.6 & 71.9 & 2.1 & 85 & 35.9 & 183 \\
\addlinespace[2pt]
OWL-TAMP~\cite{kumar2026owltamp}               & \multirow{2}{*}{17.8} & 2.0 & 12.1 & 49.1 & 6.3 & 730 & 115.3 & 790 \\
\icl                                   & & 8.0 & 14.3 & 53.3 & 6.4 & 740 & 114.6 & 803 \\
\addlinespace[2pt]
Inner Monologue~\cite{huang2023inner} & \multirow{2}{*}{64.4} & 6.0 & 43.6 & 76.5 & 14.4 & 956 & 62.3 & 1114 \\
\icl                                   & & 10.0 & 45.0 & 76.8 & 15.0 & 1047 & 68.4 & 1240 \\
\addlinespace[2pt]
EPoG$^\dagger$~\cite{yang2026integrated}            & \multirow{2}{*}{44.4} & 0.0 & 28.6 & 48.9 & 5.1 & 159 & 25.4 & 216 \\
\icl                                   & & 0.0 & 28.6 & 48.8 & 5.1 & 159 & 25.5 & 216 \\
\midrule
\textbf{Ours}                          & \multirow{2}{*}{\textbf{94.4}} & \underline{22.0} & \underline{68.6} & \underline{86.5} & 5.1 & 540 & 103.4 & 643 \\
\icl                                   & & \textbf{88.0} & \textbf{92.1} & \textbf{97.8} & 3.8 & 355 & 91.6 & 449 \\
\bottomrule
\end{tabular*}
\scriptsize
\begin{tablenotes}[flushleft]
\item Unmarked rows use zero-shot prompts. ICL examples are supplied only to grill prompts; kitchen results are shared. \textbf{Bold}: highest reported completion; \textbf{Underlined}: highest zero-shot completion. \textbf{Calls}: FM calls per trial; \textbf{Plan./Time}: mean planning/total time per trial; \textbf{Lat.}: median FM-call latency (all times in s). $^\dagger$The evaluated EPoG goal graph does not encode cooking.
\end{tablenotes}
\end{threeparttable}
\end{table}
We analyse the planner configuration and performance across task variants, then examine the contribution of each framework component. Comparisons with existing approaches and failure analysis establish the relative performance and remaining limitations.

\begin{figure}[t]
\centering
\includegraphics[
    width=\columnwidth,
    trim={0cm 3.1cm 2.5cm 0cm},
    clip
]{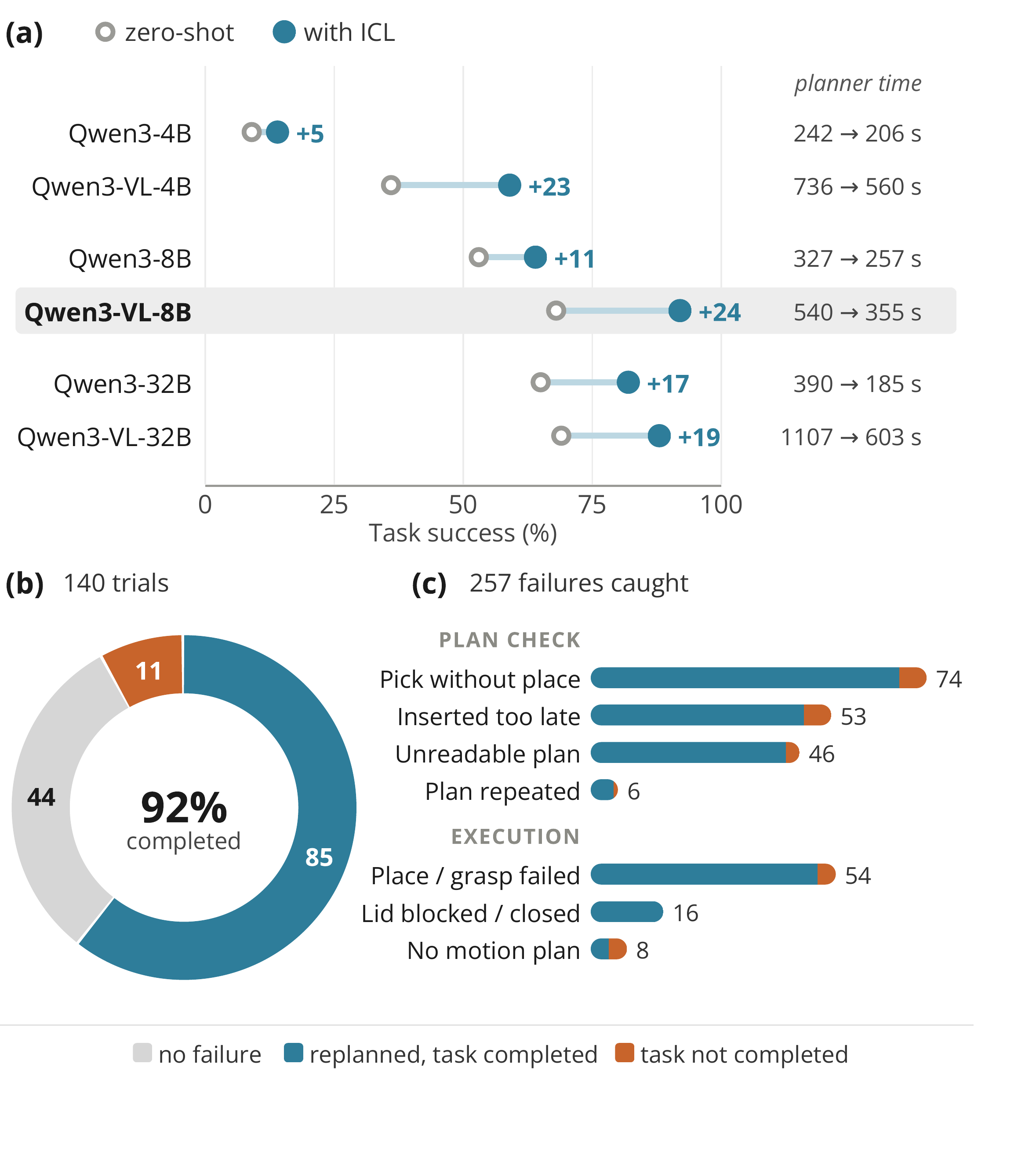}
\caption{(a) In-context example improvement across the Qwen3 (Table~\ref{tab:planner-selection}a) family.
(b) Outcomes of the selected planner over 140 trials (Table~\ref{tab:main}).
(c) ROBUST-TAMP failure analysis.}
\label{fig:failures}
\end{figure}

\noindent\phantomsection\label{RQ1}\textbf{RQ1: How do model size, input modality, and reasoning capability affect \textsc{ROBUST-TAMP}'s performance?} We first compare 4B, 8B, and 32B Qwen3-family planners with and without ICL (Table~\ref{tab:planner-selection}(a)). The 8B VLM achieves the highest success with context examples provided, improving from 68.6\% to 92.1\%. Figure~\ref{fig:failures}(a) shows that this is the largest ICL gain among the evaluated configurations, at approximately 23.6 percentage points. Its mean planning time also falls from 540 to 355\,s. Increasing the VLM size to 32B does not improve the result, reaching 87.9\% with ICL. We therefore select the 8B VLM with ICL as our planner, offering the best balance between task success and planning efficiency.

We next compare input modality and reasoning at this scale under zero-shot prompting (Table~\ref{tab:planner-selection}(b)). Qwen3-VL-8B-Thinking leads the evaluated models with 68.6\% success and 86.5\% PGC, exceeding the reasoning-enabled Qwen3-8B LLM's 53.6\% success. Within the Qwen3 families,
reasoning produces substantial improvements over both the VLM Instruct model and the LLM with thinking disabled. However, several other reasoning models perform poorly, so reasoning capability alone does not ensure executable
plans. We therefore select Qwen3-VL-8B-Thinking for the main evaluation.

With this chosen model, \textsc{ROBUST-TAMP} achieves 92.1\% success and 97.8\% PGC across the full suite (Table~\ref{tab:main}). The basic variants and both single-grocery variants reach 100\% success, covering ordinary execution, urgent obstruction removal, and deferred task completion. The phone variants reach 90\%: K1 requires clearing a placement area, whereas K2 permits the ongoing plan to continue without a discovery-triggered correction. Grill success remains at 80--90\%, where newly revealed meat must be incorporated without violating cooking and serving requirements. Increasing hidden groceries from one to three preserves 90--100\% success but raises mean trial time from 480 to 580\,s. Adding independent groceries also preserves 90\% success in the phone-obstruction variants, although waiting time does not decrease consistently. First-proposal urgency accuracy is 58.8\% overall, showing that successful task completion does not always require the FM to choose the expected urgency on its first attempt. 

\noindent\phantomsection\label{RQ2}\textbf{RQ2: How does each component contribute to task success and execution efficiency?} Table~\ref{tab:ablations} examines the contribution of individual components to the framework's overall performance. Removing memory reduces success from 95.0\% to 85.0\% in the kitchen and from 86.7\% to 60.0\% in the grill, supporting the value of retaining observations beyond the current view. Selective replanning reduces planning overhead: without \textit{IF}, kitchen planner calls rise from 4.2 to 5.5 per trial with no improvement in success, and mean planning time increases by 64\,s. The insertion ablations expose the importance of placing corrections appropriately. Always appending them reduces kitchen success to 52.5\%, where an obstruction may need clearing before a pending placement. In the grill, both fixed-front and fixed-end insertion achieve only 30.0\% success, showing that neither position suits all procedural corrections. Replacing corrective insertion with full-replan also performs worse in both domains. The dedicated \textit{WHEN} ablation isolates that scheduling change. Continuing independent actions reduces mean waiting per corrective episode from 127 to 118\,s in the kitchen and from 147 to 135\,s in the grill. Kitchen trials finish 26\,s sooner on average, whereas grill duration changes little. Concurrency therefore offers a workload-dependent timing benefit. Together, the ablations show that memory, visual input, and appropriate corrective insertion support task success, while selective replanning avoids unnecessary planner calls and concurrent execution reduces waiting.

\noindent\phantomsection\label{RQ3}\textbf{RQ3: How does \textsc{ROBUST-TAMP} compare with existing approaches with and without in-context examples?} Table~\ref{tab:baselines} compares \textsc{ROBUST-TAMP} with VLM-TAMP~\cite{yang2025guiding},
OWL-TAMP~\cite{kumar2026owltamp}, Inner Monologue~\cite{huang2023inner}, and
EPoG~\cite{yang2026integrated}. All methods use Qwen3-VL-8B-Thinking and share the same simulation perception and execution backend, while retaining their original planning interfaces. \textsc{ROBUST-TAMP} achieves the highest success and partial goal completion in both prompting settings. Without examples, it reaches 68.6\% success, compared with 43.6\% for Inner Monologue, the strongest external baseline. ICL is applied only to grill tasks, where procedural ordering is not fully specified by the primitive action descriptions. The examples demonstrate analogous ordering constraints without providing grill-specific plans; full prompts are given in the supplementary material. With ICL, \textsc{ROBUST-TAMP} reaches 92.1\% success, while Inner Monologue remains the strongest baseline at 45.0\%. The examples provide only modest gains for the other methods and no gain for EPoG. EPoG's grill results are further limited by its evaluated goal graph, which does not encode cooking. Although VLM-TAMP and EPoG are faster, they complete substantially fewer tasks; compared with Inner Monologue, \textsc{ROBUST-TAMP} also requires fewer planner calls and less planning and total trial time. Overall, it provides the strongest trade-off between task completion and planning cost.












\noindent\phantomsection\label{RQ4}\textbf{RQ4: What causes the remaining failures, and how effectively can the framework recover?}  Figure~\ref{fig:failures}(c) records 257 failure events, of which 179 are detected during plan checking. Incomplete pick--place sequences and corrections inserted too late are the largest categories. Execution failures mainly involve unsuccessful grasps, placements, and motion-planning problems. Figure~\ref{fig:failures}(b) shows that 44 trials complete without recorded failures and another 85 complete after encountering them; 11 remain unsuccessful. Thus, 88.5\% of trials with recorded failures ultimately
succeed. This recovery includes revising rejected proposals before execution and replanning after execution errors. 

\section{Conclusion}
\label{sec:discussion}
During manipulation, newly revealed objects can change what the robot must do to complete the task. We present \textsc{Robust-TAMP}, which addresses this through three connected decisions: (i) whether replanning is needed, (ii) which actions can continue during replanning, and (iii) where corrections should enter the remaining plan. Across the evaluated tasks, the \textsc{Robust-TAMP} improves task completion over four baselines. The ablations highlight the importance of scene memory and correction placement, while selective replanning and continued execution reduce unnecessary calls and waiting. The planner evaluation further shows that analogous examples improve procedural planning, whereas larger models do not consistently improve success. However, many successful trials still depend on validation and recovery, leaving reliable initial decisions as a key limitation. Future research will focus on improving urgency assessment and action ordering to reduce repeated corrections during execution.






\bibliographystyle{ACM-Reference-Format}
\bibliography{references}

\newpage
\appendix
\setcounter{section}{0}\renewcommand{\thesection}{S\arabic{section}}\renewcommand{\theHsection}{supp.\arabic{section}}
\setcounter{table}{0}\renewcommand{\thetable}{S\arabic{table}}\renewcommand{\theHtable}{supp.\arabic{table}}
\setcounter{figure}{0}\renewcommand{\thefigure}{S\arabic{figure}}\renewcommand{\theHfigure}{supp.\arabic{figure}}
\setcounter{equation}{0}\renewcommand{\theequation}{S\arabic{equation}}\renewcommand{\theHequation}{supp.\arabic{equation}}
\renewcommand{\dbltopfraction}{0.9}\renewcommand{\topfraction}{0.9}\renewcommand{\textfraction}{0.03}
\renewcommand{\dblfloatpagefraction}{0.85}\setcounter{dbltopnumber}{2}
\sloppy\raggedbottom

\section*{Supplementary Material}

\section{Task Specification and Evaluation}
\label{sec:supp-tasks}
\noindent This supplement gives implementation details, prompts and experimental settings. Notation and variant IDs
\ifsuppcombined follow Sections~\ref{sec:problem_formulation}--\ref{sec:experiments}; the variants are listed in
Table~\ref{tab:main}.\else follow the main paper; the variants are listed in its Table~3 (\emph{Task variants and ROBUST-TAMP performance}).\fi

\noindent\textbf{Goals.} Kitchen: \emph{``move ALL THE GROCERIES inside the cupboard and ALL THE MUGS inside the
box''}.\\
Grill: \emph{``COOK all raw meat using the grill and SERVE all cooked meat on the PLATE in the serving area.''}

\noindent\textbf{Placement areas.} The robot-side half of the box interior and the grill slot nearest the robot
(every meat is placed there). Non-obstructing hidden objects lie in the far half or far slots. Areas never appear
in prompts.

\noindent\textbf{Relations $\mathcal{R}_\tau$.} Evaluated on final poses over all objects, including hidden ones that
were never observed. Kitchen: every mug in the box (including the mug stored in the cupboard), every grocery on the
shelf, every phone outside the box placement area. Grill: plate in the serving area, every meat on the plate.
$|\mathcal{R}_\tau|$ is 5--8 in the kitchen and 2--4 in the grill.

\noindent\textbf{Cooking conditions $\mathcal{P}_\tau$.} One per meat, evaluated by replaying the executed history
$h_T$. A meat completes a cycle when it is inside at \texttt{close(grill\_lid)} and the lid next opens; opening the
initially closed lid completes none. Its cooking level starts at 1 if it starts cooked, else 0. The condition holds
if the level reaches 1, no cycle occurs once it is $\geq 1$ (\emph{not cooked again}), and the meat is never placed
on the plate at level 0 (\emph{not served raw}). Violations are permanent.

\noindent\textbf{SR and PGC.} $c_i=|\mathcal{R}_\tau|+|\mathcal{P}_\tau|$ (\ifsuppcombined Eq.~\ref{eq:outcome_metrics}\else main Eq.~3\fi). Only executed actions enter
$h_T$: rejected plans never execute, and a placement that fails its post-check is not recorded (relations still use
the actual pose).

\noindent\textbf{Randomization.} Seeds 0--9 perturb free objects by up to $\pm3$\,cm in $x,y$ and $\pm20^\circ$ in yaw.
A sample is rejected if it touches another movable object or its height changes by more than 1\,cm after settling;
after 30 rejections the nominal pose is kept.

\section{Implementation Details}
\label{sec:supp-impl}

\subsection{Perception and Memory}
\label{sec:supp-perception}

At observation $t$, visible objects $V_t$ store their current region and observation index in memory $m_t$.
Objects that leave view keep their latest entry; unseen objects have no entry and are not passed to the planner.
Let $\mathrm{region}_t(o)$ be the observed region of $o\in V_t$. Starting from $m_{-1}=\emptyset$,
\begin{equation}
\label{eq:supp-memory}
m_t(o)=
\begin{cases}
\big(\mathrm{region}_t(o),t\big), & o\in V_t,\\
m_{t-1}(o), & o\in \mathrm{dom}(m_{t-1})\setminus V_t.
\end{cases}
\end{equation}
Equation~\ref{eq:supp-memory} simply overwrites the memory entry of a visible object and keeps the last entry for an object that is no longer visible. Thus, $\mathrm{dom}(m_t)=\mathrm{dom}(m_{t-1})\cup V_t$.
Observations are taken after each execution bundle and before each FM request; validation re-queries reuse the latest observation.

An object is visible if its segmentation mask covers at least 10 pixels (50 in the grill), with $V_t$ taken over five cameras. Lid state is read separately from articulation state. Regions are assigned from padded region boxes; at least 35\% overlap with the box interior assigns an object to the box. The horizontal object area $\mathrm{area}(o)$ is retained from the latest observation for placement checks. Remembered objects are listed separately in prompts with their last region and time since observation. Under $-$Memory, prompts and \textit{IF} use only $V_t$.

\subsection{Replanning, Concurrency and Insertion}
\label{sec:supp-replanning}

\noindent\textbf{\textit{IF}: does an object require correction?}
After each completed bundle, the framework checks every visible or remembered object. A correction is requested only if the object is not already covered by $\pi_t$ and either (i) it is task-relevant but incomplete, or (ii) it is task-irrelevant but blocks a pending placement. A pending \texttt{pick}/\texttt{place} covers the object. An accepted \texttt{NO\_ACTIONS} response also covers it until its recorded region changes. Task completion may depend on $h_t$; for example, meat on the plate is complete only if it was cooked.

Formally,
\begin{equation}
\label{eq:supp-trigger}
O_t^{\mathrm{trig}}
=
\left\{
 o\in\mathrm{dom}(m_t)
 \;\middle|\;
 \begin{aligned}
 &\neg\,\mathrm{covered}(o)\;\wedge\\[-1mm]
 &\big[(\mathrm{task}(o)\wedge\neg\,\mathrm{done}(o))\\[-1mm]
 &\qquad\vee(\neg\,\mathrm{task}(o)\wedge\mathrm{blocks}(o))\big]
 \end{aligned}
\right\}.
\end{equation}
Equation~\ref{eq:supp-trigger} is the exact \textit{IF} test: an object enters the trigger set only when it is not already handled and either still belongs to the task or blocks a required placement. Here, $\mathrm{task}(o)$ marks task-relevant categories, $\mathrm{done}(o)$ checks the goal region and cooking history when needed, and $\mathrm{blocks}(o)$ checks overlap with any pending placement area. All objects in $O_t^{\mathrm{trig}}$ form one event $e_t$; an empty set causes no FM call. Under $-$IF, every newly visible object triggers.

\noindent\textbf{\textit{WHEN}: which bundles may continue?}
When \textit{IF} raises $e_t$, the FM runs in the background while independent bundles continue. The scheduler reserves the triggering objects, their goal regions, obstructed container/staging regions, and associated lids, then scans $\pi_t$ in order. A bundle may execute only if its object, source region, and destination region do not use reserved resources. If a bundle is withheld, its resources are reserved too, so dependent later bundles also wait. Eligible bundles preserve their relative order. Their anticipated effects are not added to $s_t^{\mathrm{vis}}$ before execution. If a scheduled bundle fails, the concurrent FM response is discarded. Under $-$WHEN, the request is synchronous.

For bundle $b$, let $\mathrm{obj}(b)$, $\mathrm{src}(b)$, and $\mathrm{dst}(b)$ denote its manipulated object, source region, and destination region. The scheduler keeps three reservation sets: $\mathcal{B}_{\mathrm{obj}}$ for objects, $\mathcal{B}_{\mathrm{dst}}$ for destinations, and $\mathcal{B}_{\mathrm{src}}$ for sources. Initially,
\begin{equation}
\label{eq:supp-reservations}
\begin{aligned}
\mathcal{B}_{\mathrm{dst}}
&= \{\mathrm{goal}(o):o\in O_t^{\mathrm{trig}}\}\\
&\quad\cup\mathcal{S}\cup\mathcal{C},\\
\mathcal{B}_{\mathrm{src}}
&= \mathcal{B}_{\mathrm{dst}}\setminus\mathcal{S},\\
\mathcal{B}_{\mathrm{obj}}
&= O_t^{\mathrm{trig}}\cup\mathcal{L}_{\mathrm{res}}.
\end{aligned}
\end{equation}
Equation~\ref{eq:supp-reservations} reserves the trigger objects, their required destination regions, staging/container regions, and associated lids before concurrent execution is considered.

A bundle is eligible iff
\begin{equation}
\label{eq:supp-eligible}
\begin{aligned}
\mathrm{obj}(b)\cap\mathcal{B}_{\mathrm{obj}} &= \emptyset,\\
\mathrm{dst}(b)\cap\mathcal{B}_{\mathrm{dst}} &= \emptyset,\\
\mathrm{src}(b)\cap\mathcal{B}_{\mathrm{src}} &= \emptyset.
\end{aligned}
\end{equation}
Equation~\ref{eq:supp-eligible} therefore allows $b$ to run only when its object, destination, and source are all independent of the reserved resources. If $b$ is withheld, its object and used regions are added to the corresponding sets so dependent later bundles also wait. Placements into staging always wait; picks from staging wait only after a withheld placement into that region.

\noindent\textbf{\textit{WHERE}: where does the correction enter?}
The FM response $r_t$ contains corrective blocks with an urgency and insertion position: \texttt{front}, \texttt{after} a pending action, or \texttt{end}. Actions completed during inference are first removed from $\pi_t$. Urgent blocks must use \texttt{front}; anchored blocks are inserted after a still-pending, unscheduled action; end blocks are appended. Existing pending actions keep their relative order. Accepted \texttt{NO\_ACTIONS} blocks suppress another \textit{IF} trigger until the object's recorded region changes.

Let $\pi_t^{-}$ be the plan left after actions completed during inference are removed, $F_1,\ldots,F_k$ the front blocks, $A$ the anchored blocks, and $Z_1,\ldots,Z_\ell$ the end blocks. The merge is
\begin{equation}
\label{eq:supp-merge}
\begin{aligned}
\pi_t \leftarrow {}& F_1\cdots F_k\\
&\cdot\mathrm{ins}(\pi_t^{-},A)\\
&\cdot Z_1\cdots Z_\ell.
\end{aligned}
\end{equation}
Equation~\ref{eq:supp-merge} keeps the unfinished original plan in order, places front blocks before it, inserts anchored blocks after their specified pending actions, and appends end blocks. The implementation technically allows a deferred block at \texttt{front}; this occurred in 4 of 294 accepted action blocks. Fixed-front and fixed-end ablations override the returned position.

\noindent\textbf{Validation.}
Before merging, the full correction is checked using Table~\ref{tab:supp-validation}. Any failure rejects the response and triggers a re-query. The validator checks syntax, coverage, action structure, insertion validity, and placement order; procedural urgency itself remains an FM decision.

\begin{table}[t]
\centering
\footnotesize
\setlength{\tabcolsep}{3pt}
\renewcommand{\arraystretch}{1.05}
\caption{Checks applied to a corrective response before merging.}
\label{tab:supp-validation}
\begin{tabularx}{\columnwidth}{@{}l>{\raggedright\arraybackslash}X@{}}
\toprule
Category & Rejected if \\
\midrule
Format & a block is malformed or has missing/invalid fields \\
Coverage & a trigger object is omitted, or an unrelated object is included \\
References & an action/object/region is unsupported or unknown \\
Manipulation & the block contains an invalid or incomplete manipulation sequence \\
Insertion & an urgent block is not at \texttt{front}, or its anchor is invalid \\
Merged plan & the merged plan fails re-parsing from the current executor state \\
Placement order & a blocking object is removed only after the placement it blocks \\
\bottomrule
\end{tabularx}
\end{table}

\subsection{Recovery and Termination}
\label{sec:supp-termination}

These mechanisms are separate from \textit{IF}, \textit{WHEN}, and \textit{WHERE}.
\begin{itemize}\setlength{\itemsep}{0pt}\setlength{\parskip}{0pt}
    \item \emph{Validation re-query.} Rejected plans are re-queried with the current scene, completed actions, and one feedback line; the remaining plan is preserved.
    \item \emph{Execution-failure replanning.} A failed grasp, placement, motion plan, or pre-action check requests a replacement of the remaining plan from the updated scene and history.
    \item \emph{Primitive retries.} Pose resampling, motion attempts, release retries, and grasp-attachment retries occur inside a primitive and do not call the FM.
    \item \emph{Termination.} A non-replannable executor failure, repeated identical output after one retry, or plan exhaustion with no new trigger ends the trial.
    \item \emph{Call budget.} At most 11 FM calls are allowed per trial: one initial request and ten further calls, including re-queries and discarded concurrent calls. Baselines use their own limits (Section~\ref{sec:supp-baselines}).
\end{itemize}
Termination does not imply success; success is evaluated afterwards (Section~\ref{sec:supp-tasks}).

\subsection{Simulation Controllers}
\label{sec:supp-controllers}

A Franka Panda runs in MuJoCo with damped-least-squares IK and joint-space RRT-Connect. Kitchen bundles outside the cupboard use PDDLStream (stable-pose, grasp/IK, and motion streams; $\leq60$\,s per solve). Box placements are sampled inside the placement area, while cupboard objects use a scripted approach. The box lid uses a recorded opening near its nominal pose and planning otherwise. Grill transfers are planned in stages; meat uses a fixed grill pose, plate/lid opening use recorded trajectories with a runtime fallback, and closing drives the hinge. Pre/post-checks detect blocked access, dropped objects, insufficient opening, and off-target placements.


\section{Prompts and Recorded Examples}
\label{sec:supp-prompts}
\newcommand{\paneltitle}[2]{\textsc{#1}\,\textperiodcentered\,#2}
\tcbset{tight/.style={unbreakable, left=1mm, right=1mm, top=2.2mm, bottom=0.2mm, before skip=2mm, after skip=0.6mm,
  fonttitle=\sffamily\bfseries\scriptsize, boxed title style={left=1mm, right=1mm, top=0.2mm, bottom=0.2mm},
  attach boxed title to top left={yshift=-1.9mm, xshift=2mm}}}
\tcbset{notebox/.style={enhanced, unbreakable, colback=pbGray!4!white, colframe=pbGray!70!black, boxrule=0.5pt,
  arc=1.5mm, left=1mm, right=1mm, top=2.2mm, bottom=0.6mm, fontupper=\scriptsize,
  fonttitle=\sffamily\bfseries\scriptsize, coltitle=white, before skip=2mm, after skip=0.6mm,
  attach boxed title to top left={yshift=-1.9mm, xshift=2mm},
  boxed title style={colback=pbGray!70!black, arc=1mm, boxrule=0pt, left=1mm, right=1mm, top=0.2mm, bottom=0.2mm}}}
\newcommand{\exrow}[1]{\par\vspace{0.3mm}\noindent{\sffamily\bfseries\footnotesize #1}\par\nobreak\vspace{-0.8mm}}

\begin{figure*}[t]
\noindent\begin{minipage}[t]{0.595\textwidth}\vspace{0pt}
\begin{promptbox}[tight]{pbBlue}{\paneltitle{Input}{System message, grill (verbatim)}}
You are the task planner for a robot arm. Given a goal and the current state of a scene, you output the ordered list of actions that achieves the goal.

Actions available in this scene:
pick(o): grasp object o.
  Preconditions: the gripper is empty; o is a listed object; o is not in a region closed off by a closed lid.
  Effects: the gripper holds o.
place(o, r): put the held object o into region r.
  Preconditions: the gripper holds o; r is a listed region; r is not closed off by a closed lid.
  Effects: o is in r; the gripper is empty.
open(l): open lid l.
  Preconditions: the gripper is empty; l is closed; no object is on the top surface of l.
  Effects: l is open; the regions l closes off become reachable and their contents can become visible.
close(l): close lid l.
  Preconditions: the gripper is empty; l is open.
  Effects: l is closed; the regions l closes off become unreachable and their contents are no longer visible.

Output format:
Reasoning is allowed. End your answer with a line containing only FINAL ACTIONS: followed by one action per line, in execution order, in this form:
FINAL ACTIONS:
pick(o)
place(o, r)
Use only the action names available in this scene and the object, lid and region names given in the state. Write nothing after the last action line. If the goal is already satisfied, write NO_ACTIONS as the only line after FINAL ACTIONS:.
\end{promptbox}
\end{minipage}\hfill
\begin{minipage}[t]{0.395\textwidth}\vspace{0pt}
\begin{promptbox}[tight]{pbBlue}{\paneltitle{Input}{User message: initial request (verbatim)}}
## Goal
COOK all raw meat using the grill and SERVE all cooked meat on the PLATE in the serving area.

## Current state
Visible objects and the region each one is in:
- plate: dish_rack
- raw_meat_1: grill_side_area
Remembered objects (not currently visible): last region, steps since last seen:
- (none)
Lids:
- grill_lid: closed (closes off inside_grill)
Gripper: empty
Regions: table, grill_side_area, inside_grill, plate_top, serving_area, dish_rack
\end{promptbox}
\begin{promptbox}[tight]{pbBlue}{\paneltitle{Input}{ICL only: system message continues}}
[... format paragraph above ...]

Examples (from different scenes; their objects, lids and regions are not in your scene):

Example 1 (a request for a full plan)
[... Examples 1-3 as in Figure S2 ...]
\end{promptbox}
\end{minipage}
\begin{promptbox}[tight]{pbBlue}{\paneltitle{Input}{Corrective requests: format paragraph replacing the one above (verbatim)}}
Output format for this request:
Reasoning is allowed. End your answer with a line containing only FINAL BLOCKS: followed by one or more blocks. Each block handles some of the objects listed under "Why a new plan is requested" and has exactly this form:
BLOCK
objects: <names of the listed objects this block handles, comma-separated>
urgency: urgent or deferred
insert: front, or after <action id from the remaining plan>, or end
reason: <one short sentence>
actions:
<one action per line, in execution order>
END BLOCK
Every listed object must be handled by at least one block. An urgent block must use insert: front. The actions of a block must end with the gripper empty. If a listed object needs no action, handle it with a block whose actions section is the single line NO_ACTIONS; such a block needs no urgency or insert line. If no listed object needs any action, write NO_ACTIONS as the only line after FINAL BLOCKS:. Write nothing after the last END BLOCK.
\end{promptbox}
\Description{Text panels with the planner's system message, the initial user message and the corrective output format.}
\caption{Shared prompt (initial request of G2, seed~0, ICL run). Both conditions share the system message up to
its format paragraph; grill requests of the ICL condition continue with Figure~\ref{fig:supp-icl}. Bracketed lines
are editorial.}
\label{fig:supp-prompt}
\end{figure*}
\begin{figure*}[t]
\exrow{Example 1 (a request for a full plan) \textmd{\textendash{} laundry: complete the processing cycle before the final placement}}
\noindent\begin{minipage}[t]{0.46\textwidth}\vspace{0pt}
\begin{promptbox}[tight, equal height group=ex1]{pbBlue}{Request (verbatim)}
## Goal
DRY all wet towels using the dryer and PUT all dry towels in the BASKET in the laundry area.
## Current state
Visible objects and the region each one is in:
- basket: shelf
- wet_towel_1: counter
Lids:
- dryer_door: closed (closes off inside_dryer)
Gripper: empty
Regions: counter, shelf, inside_dryer, basket_top, laundry_area
\end{promptbox}
\end{minipage}\hfill
\begin{minipage}[t]{0.265\textwidth}\vspace{0pt}
\begin{promptbox}[tight, equal height group=ex1]{pbGreen}{Good answer (verbatim)}
FINAL ACTIONS:
pick(basket)
place(basket, laundry_area)
open(dryer_door)
pick(wet_towel_1)
place(wet_towel_1, inside_dryer)
close(dryer_door)
open(dryer_door)
pick(wet_towel_1)
place(wet_towel_1, basket_top)
Outcome: the towel came out dry and is in the basket in the laundry area. Goal met.
\end{promptbox}
\end{minipage}\hfill
\begin{minipage}[t]{0.265\textwidth}\vspace{0pt}
\begin{promptbox}[tight, equal height group=ex1]{pbRed}{Bad answer (verbatim)}
FINAL ACTIONS:
open(dryer_door)
pick(wet_towel_1)
place(wet_towel_1, inside_dryer)
pick(wet_towel_1)
place(wet_towel_1, basket_top)
Outcome: the towel is still wet, and the basket is still on the shelf. Goal not met.
\end{promptbox}
\end{minipage}
\exrow{Example 2 (a corrective request) \textmd{\textendash{} laundry: remove an already processed object before another cycle}}
\noindent\begin{minipage}[t]{0.46\textwidth}\vspace{0pt}
\begin{promptbox}[tight, equal height group=ex2]{pbBlue}{Request (excerpt: goal and state omitted)}
## Completed actions
- a1: pick(basket)
- a2: place(basket, laundry_area)
- a3: open(dryer_door)
## Remaining plan (not executed yet)
- a4: pick(wet_towel_1)
- a5: place(wet_towel_1, inside_dryer)
- a6: close(dryer_door)
- a7: open(dryer_door)
- a8: pick(wet_towel_1)
- a9: place(wet_towel_1, basket_top)
## Why a new plan is requested
- After open(dryer_door), these observed objects are not handled by the remaining plan:
  - dry_towel_2 (in inside_dryer): relevant to the goal, not in its goal state
\end{promptbox}
\end{minipage}\hfill
\begin{minipage}[t]{0.265\textwidth}\vspace{0pt}
\begin{promptbox}[tight, equal height group=ex2]{pbGreen}{Good answer (verbatim)}
FINAL BLOCKS:
BLOCK
objects: dry_towel_2
urgency: urgent
insert: front
reason: dry_towel_2 is already dry.
actions:
pick(dry_towel_2)
place(dry_towel_2, basket_top)
END BLOCK
Outcome: both towels are dry and in the basket. Goal met.
\end{promptbox}
\end{minipage}\hfill
\begin{minipage}[t]{0.265\textwidth}\vspace{0pt}
\begin{promptbox}[tight, equal height group=ex2]{pbRed}{Bad answer (verbatim)}
FINAL BLOCKS:
BLOCK
objects: dry_towel_2
urgency: deferred
insert: end
reason: dry_towel_2 can be moved at the end.
actions:
pick(dry_towel_2)
place(dry_towel_2, basket_top)
END BLOCK
Outcome: dry_towel_2 was still inside during a6 and a7 and came out scorched. Goal not met.
\end{promptbox}
\end{minipage}
\exrow{Example 3 (a corrective request) \textmd{\textendash{} germination: defer the removal until the scheduled cycle has run}}
\noindent\begin{minipage}[t]{0.46\textwidth}\vspace{0pt}
\begin{promptbox}[tight, equal height group=ex3]{pbBlue}{Request (excerpt: state omitted)}
## Goal
GERMINATE all seed trays using the germination chamber and PUT all sprouted trays on the NURSERY BENCH.
## Completed actions
- a1: open(chamber_door)
## Scheduled to run before your corrective block is applied (not executed yet)
- a2: pick(seed_tray_1)
- a3: place(seed_tray_1, inside_chamber)
- These actions run while you plan. Do not repeat them, and do not use their ids as an insertion point; the Current state does not include their effects yet.
## Remaining plan (not executed yet)
- a4: close(chamber_door)
- a5: open(chamber_door)
- a6: pick(seed_tray_1)
- a7: place(seed_tray_1, nursery_bench)
## Why a new plan is requested
- After open(chamber_door), these observed objects are not handled by the remaining plan:
  - seed_tray_2 (in inside_chamber): relevant to the goal, not in its goal state; does not lie where the remaining plan places objects
\end{promptbox}
\end{minipage}\hfill
\begin{minipage}[t]{0.265\textwidth}\vspace{0pt}
\begin{promptbox}[tight, equal height group=ex3]{pbGreen}{Good answer (verbatim)}
FINAL BLOCKS:
BLOCK
objects: seed_tray_2
urgency: deferred
insert: after a5
reason: seed_tray_2 has not germinated yet.
actions:
pick(seed_tray_2)
place(seed_tray_2, nursery_bench)
END BLOCK
Outcome: both trays sprouted during a4 and a5 and are on the nursery bench. Goal met.
\end{promptbox}
\end{minipage}\hfill
\begin{minipage}[t]{0.265\textwidth}\vspace{0pt}
\begin{promptbox}[tight, equal height group=ex3]{pbRed}{Bad answer (verbatim)}
FINAL BLOCKS:
BLOCK
objects: seed_tray_2
urgency: urgent
insert: front
reason: seed_tray_2 must be moved before the door closes.
actions:
pick(seed_tray_2)
place(seed_tray_2, nursery_bench)
END BLOCK
Outcome: seed_tray_2 was taken out before a4 and a5 and never sprouted. Goal not met.
\end{promptbox}
\end{minipage}
\Description{Three rows of text panels, each with an example request, a good answer and a bad answer.}
\caption{In-context examples of our ICL condition; the baselines received Examples~1 and~2 in their own formats.}
\label{fig:supp-icl}
\end{figure*}
\begin{figure*}[t]
{\sffamily\bfseries\footnotesize Accepted correction (G2, seed~0)}\par
\noindent\begin{minipage}[t]{0.575\textwidth}\vspace{0pt}
\begin{promptbox}[tight]{pbBlue}{\paneltitle{Input}{User message text (excerpt: goal and current state omitted)}}
## Completed actions
- a1: pick(plate)
- a2: place(plate, serving_area)
- a3: open(grill_lid)

## Scheduled to run before your corrective block is applied (not executed yet)
- a4: pick(raw_meat_1)
- a5: place(raw_meat_1, inside_grill)
- These actions run while you plan. Do not repeat them, and do not use their ids as an insertion point; the Current state does not include their effects yet.

## Remaining plan (not executed yet)
- a6: close(grill_lid)
- a7: open(grill_lid)
- a8: pick(raw_meat_1)
- a9: place(raw_meat_1, plate_top)

## Why a new plan is requested
- After open(grill_lid), these observed objects are not handled by the remaining plan:
  - cooked_meat_1 (in inside_grill): relevant to the goal, not in its goal state; does not lie where the remaining plan places objects
  - raw_meat_2 (in inside_grill): relevant to the goal, not in its goal state; does not lie where the remaining plan places objects

## What to plan
- Plan actions only for the objects listed under "Why a new plan is requested". The remaining plan stays as it is: do not repeat, reorder or remove its actions.
- Do not move objects that are already in their goal state, and do not move objects that are not relevant to the goal unless they are listed.
- Decide each block's urgency and where it goes in the remaining plan by considering what would go wrong if its actions were delayed.
\end{promptbox}
\end{minipage}\hfill
\begin{minipage}[t]{0.415\textwidth}\vspace{0pt}
\begin{promptbox}[tight]{pbGreen}{\paneltitle{Output}{Answer after reasoning (verbatim)}}
FINAL BLOCKS:
BLOCK
objects: cooked_meat_1
urgency: urgent
insert: front
reason: it is already cooked but not served on the plate.
actions:
pick(cooked_meat_1)
place(cooked_meat_1, plate_top)
END BLOCK
BLOCK
objects: raw_meat_2
urgency: deferred
insert: after a7
reason: it needs to be cooked before serving, which occurs after the lid is opened.
actions:
pick(raw_meat_2)
place(raw_meat_2, plate_top)
END BLOCK
FINAL BLOCKS:
\end{promptbox}
\begin{tcolorbox}[notebox, title={Merged remaining plan $\widehat{\pi}$ (logged)}]
{\ttfamily\scriptsize a10 pick(cooked\_meat\_1)\\ a11 place(cooked\_meat\_1, plate\_top)\\ a6 close(grill\_lid)\quad a7 open(grill\_lid)\\
a12 pick(raw\_meat\_2)\\ a13 place(raw\_meat\_2, plate\_top)\\ a8 pick(raw\_meat\_1)\\ a9 place(raw\_meat\_1, plate\_top)}
\end{tcolorbox}
\end{minipage}

\vspace{1mm}{\sffamily\bfseries\footnotesize Rejected correction and re-query (G1-n1, seed~2)}\par
\noindent\begin{minipage}[t]{0.325\textwidth}\vspace{0pt}
\begin{promptbox}[tight, equal height group=rej]{pbGreen}{\paneltitle{Output}{Rejected answer (verbatim)}}
FINAL BLOCKS:
BLOCK
objects: cooked_meat_1
urgency: urgent
insert: after a5
reason: cooked_meat_1 is already cooked and needs to be served before closing the grill
actions:
pick(cooked_meat_1)
place(cooked_meat_1, plate_top)
END BLOCK
\end{promptbox}
\end{minipage}\hfill
\begin{minipage}[t]{0.33\textwidth}\vspace{0pt}
\begin{promptbox}[tight, equal height group=rej]{pbRed}{\paneltitle{Feedback}{Line added to the re-query (verbatim)}}
- Your previous blocks were rejected (failure code invalid_corrective_block): block 1 inserts after a5, which is not in the remaining plan
\end{promptbox}
\end{minipage}\hfill
\begin{minipage}[t]{0.325\textwidth}\vspace{0pt}
\begin{promptbox}[tight, equal height group=rej]{pbGreen}{\paneltitle{Output}{Accepted revision (verbatim)}}
FINAL BLOCKS:
BLOCK
objects: cooked_meat_1
urgency: urgent
insert: front
reason: cooked_meat_1 must be served before closing the grill lid to prevent it from being inaccessible.
actions:
pick(cooked_meat_1)
place(cooked_meat_1, plate_top)
END BLOCK
\end{promptbox}
\end{minipage}
\Description{Text panels with a recorded corrective request, the model's answers, a feedback line and the merged plan.}
\caption{Recorded corrective requests of the ICL run. Top: an accepted correction and its logged merge. Bottom: a
rejected answer, the feedback line added to the re-query, and the accepted revision.}
\label{fig:supp-correction}
\end{figure*}

\noindent Figures~\ref{fig:supp-prompt}--\ref{fig:supp-correction} show recorded requests and answers of our
evaluated ICL run (Qwen3-VL-8B-Thinking). Panel titles mark each panel: \emph{verbatim} (unchanged recorded text),
\emph{excerpt} (recorded text with the named sections omitted) or editorial (bracketed lines, our notes). Answers are
the logged text after the model's reasoning; camera images are not shown.

\noindent\textbf{Requests.} A system message (role, actions with preconditions and effects, output format; kitchen:
\texttt{pick}, \texttt{place}, \texttt{open}; grill adds \texttt{close}) and a user message (goal, state; replans
add completed actions, the remaining plan with identifiers and the reason). Corrective requests swap in the block
format, list actions scheduled during inference, and end with \emph{What to plan}.
\label{sec:supp-prompt-zs}

\noindent\textbf{In-context examples.} Supplied only in the ICL condition: ROBUST-TAMP receives Examples~1--3
(Figure~\ref{fig:supp-icl}), appended to the system message of every grill request; the baselines receive Examples~1
and~2 only, in their own formats (Section~\ref{sec:supp-baselines}). Kitchen requests are identical in both conditions.
The examples come from other scenes, give no grill-specific instruction, and demonstrate the procedure through good
and bad answers: procedural completion, i.e.\ a full cycle before the final placement (Ex.~1); an urgent correction,
removing an already processed object before another cycle (Ex.~2); and a deferred correction, leaving an unprocessed
object until the scheduled cycle has run (Ex.~3).
\label{sec:supp-icl}

\noindent\textbf{Recorded corrections} (Figure~\ref{fig:supp-correction}). \emph{G2, seed~0}: three FM calls; the
initial plan failed the plan check (open while holding \texttt{raw\_meat\_1}) and its re-query passed. Opening the
grill revealed a cooked and a raw meat; during the 142\,s corrective call the robot ran the scheduled \texttt{a4}--\texttt{a5}
(7.5\,s), leaving \texttt{a6}--\texttt{a9} as anchors. The correction was accepted first time and the trial succeeded.
\emph{G1-n1, seed~2}: the answer anchored on the scheduled \texttt{a5} (and paired \texttt{urgent} with a non-front
position); the re-query (\texttt{a1}--\texttt{a5} completed, one feedback line) was accepted and the trial succeeded.
\label{sec:supp-correction}

\section{Baseline Adaptations}
\label{sec:supp-baselines}
\noindent All baselines run inside our pipeline (same robot, executor and checks, variants, seeds, evaluator and
planner model) with our memory, \textit{IF}, concurrency and insertion disabled. Requests carry the visible objects
and regions, lid and gripper states, the region list and, for VLMs, an image. The executor lets a baseline pick any
previously observed object, but no baseline receives our memory entries; what each retains is stated below. Every
model completion counts as a call.

\noindent\textbf{VLM-TAMP.} Two calls per round (English subgoals, then formal subgoals); a round repeats only when a
subgoal fails, at most twice ($\leq 6$ calls). Re-prompts add achieved subgoals, gripper content, the failed subgoal and
colliding bodies. Seen objects are kept at their last region and listed with the visible ones; revealed objects reach
the model only through a re-prompt.

\noindent\textbf{OWL-TAMP.} All calls precede execution and use the initial observation: one plan sketch, one
goal-constraint call and one constraint call per placement operator (no cap). Search with up to five skeletons, then
open-loop execution: hidden objects never enter the plan.

\noindent\textbf{Inner Monologue.} Queried after every bundle and every unusable answer, with a success line and
newly visible objects (or none). The dialogue replaces the original few-shot prompts inside our zero-shot structure,
without a remembered-objects section. Ends after more than ten failures or 30 queries.

\noindent\textbf{EPoG.} A goal graph maps each observed object's category to its goal region, without a model call or
unseen-object estimation; new objects trigger a model-free replan ($\leq 10$). The model is called only to resolve a
simulated motion error ($\leq 10$ calls). The graph holds final relations only, so cooking is not represented.

\noindent\textbf{ICL.} Baselines received Examples~1 and~2 (Figure~\ref{fig:supp-icl}) in grill requests only, in
their own formats: VLM-TAMP subgoal answers (first query of each round), OWL-TAMP operator plans (sketch request),
Inner Monologue monologue answers (system message), EPoG error-resolution examples (resolve request). EPoG's model was
called 0.3 times per grill trial, so its examples rarely reached it.

\section{Experimental Configuration}
\label{sec:supp-config}
\begin{table}[t]
\centering
\footnotesize
\setlength{\tabcolsep}{3pt}
\renewcommand{\arraystretch}{1.05}
\caption{Planner inference settings (logged with every trial and in the run records).}
\label{tab:supp-inference}
\begin{tabularx}{\columnwidth}{@{}l>{\raggedright\arraybackslash}Y@{}}
\toprule
Model & Qwen3-VL-8B-Thinking (BF16), reasoning on \\
Sampling & temperature 1.0, top-$p$ 0.95, top-$k$ 20, presence penalty 0 \\
Output limit & 24{,}576 tokens per call \\
Visual input & one composite image per request \\
Engine & vLLM 0.30.0, one server per run, context 32{,}768, GPU memory 0.90, Qwen3 reasoning parser, $\leq 2$ images \\
Timeout & 1{,}800\,s per call \\
Hardware & NVIDIA RTX 5090, 32\,GB \\
\bottomrule
\end{tabularx}
\end{table}

\noindent\textbf{Runs.} Ten trials per variant (seeds 0--9); ICL runs execute the 50 grill trials and share the
zero-shot kitchen trials. Our ICL run differs from zero-shot only in the examples and one parser change (the last
\texttt{FINAL} marker followed by content is used). Baseline zero-shot and ICL runs differ only in the examples.

\noindent\textbf{Concurrency and timing.} Six trials run concurrently against one server, so all timings include this
load; baseline runs were executed one after another on the same server and GPU. Trials stop after 14{,}400\,s.
Infrastructure failures (no final evaluation) are rerun at most twice; none occurred, and the longest trial took
3{,}981\,s. Call budgets: Sections~\ref{sec:supp-termination} and~\ref{sec:supp-baselines}.

\noindent\textbf{Idle.} Per discovery-triggered episode (trigger call plus its re-queries): the trigger call's latency
minus the time spent executing scheduled bundles (full latency under $-$WHEN), plus each re-query's latency;
averaged over all episodes. Initial planning and failure replans are excluded.

\noindent\textbf{Ablations.} K1--K4 zero-shot, G1--G3 with the examples of Figure~\ref{fig:supp-icl}; full-system rows
reuse the main runs. $-$WHERE also disables concurrency, and the LLM planner (Qwen3-8B) receives no image.

\noindent\textbf{Real-world demonstrations.} The five real-world settings use the robot setup shown
in Figure~1 of the main paper. Their corresponding execution videos are provided in the
\href{https://narendhiranv04.github.io/ROBUST-TAMP/}
{\textit{project video gallery}}.


\end{document}